\documentclass[10pt,twocolumn,letterpaper]{article}

\usepackage{cvpr}              % To produce the CAMERA-READY version

\usepackage[accsupp]{axessibility}  % Improves PDF readability for those with disabilities.

\usepackage{graphicx}
\usepackage{amsmath}
\usepackage{amssymb}
\usepackage{booktabs}
\usepackage{color}
\usepackage{url}
\usepackage{cite}
\usepackage[ruled,vlined,linesnumbered]{algorithm2e}
\usepackage{algpseudocode}
\usepackage{amsmath, amscd, amsthm, amssymb}
\usepackage{array}
\usepackage{caption}
\usepackage{subcaption}
\usepackage{multirow}
\usepackage{diagbox}
\usepackage{lipsum}

\usepackage{enumitem}

\newif\ifshowcomment
\showcommenttrue
\ifshowcomment
    \newcommand{\qz}[1]{[{\color{cyan}Qingzhao: #1}]}
    \newcommand{\st}[1]{[{\color{blue}Shengtuo: #1}]}
    \newcommand{\jc}[1]{[{\color{red}Jiachen: #1}]}
\else
    \newcommand{\qz}[1]{}
    \newcommand{\st}[1]{}
    \newcommand{\jc}[1]{}
\fi

\usepackage[pagebackref,breaklinks,colorlinks]{hyperref}

\usepackage[capitalize]{cleveref}
\crefname{section}{Sec.}{Secs.}
\Crefname{section}{Section}{Sections}
\Crefname{table}{Table}{Tables}
\crefname{table}{Tab.}{Tabs.}

\def\cvprPaperID{4282} % *** Enter the CVPR Paper ID here
\def\confName{CVPR}
\def\confYear{2022}

\begin{document}

\title{On Adversarial Robustness of Trajectory Prediction for Autonomous Vehicles}

\author{
Qingzhao Zhang$^1$, Shengtuo Hu$^1$, Jiachen Sun$^1$, Qi Alfred Chen$^2$, Z. Morley Mao$^1$\\
$^1$University of Michigan, $^2$University of California, Irvine \\
{\tt\small \{qzzhang, shengtuo, jiachens, zmao\}@umich.edu \, alfchen@uci.edu}
}

\maketitle
\begin{abstract}
Trajectory prediction is a critical component for autonomous vehicles (AVs) to perform safe planning and navigation.
However, few studies have analyzed the adversarial robustness of trajectory prediction or investigated whether the worst-case prediction can still lead to safe planning.
To bridge this gap, we study the adversarial robustness of trajectory prediction models by proposing a new adversarial attack that perturbs normal vehicle trajectories to maximize the prediction error.
Our experiments on three models and three datasets show that the adversarial prediction increases the prediction error by more than 150\%.
Our case studies show that if an adversary drives a vehicle close to the target AV following the adversarial trajectory, the AV may make an inaccurate prediction and even make unsafe driving decisions.
We also explore possible mitigation techniques via data augmentation and trajectory smoothing.
\end{abstract}

\vspace{-1em}
\section{Introduction}

Autonomous vehicles (AVs) are transforming the transportation systems. 
AV is a complex system integrating a pipeline of modules such as perception of obstacles, planning of driving behaviors, and controlling of the physical vehicle~\cite{apollo, autoware}.
Specifically, trajectory prediction in the perception module predicts the future trajectories of nearby moving objects. The prediction is essential for the planning module and affects AV's driving behavior. Therefore, accurate trajectory prediction is critical for safe AV driving.
% For instance, when AV is about to do an unprotected right turn at an intersection, if it correctly predicts that a car from the left is driving through, the AV's planning module will yield the right-of-way according to the driving rules.
% Hence, accurate trajectory prediction is critical for safe AV driving.

Many studies propose trajectory prediction models based on deep neural networks~\cite{chandra2020forecasting,yuan2021agentformer,phan2020covernet,deo2020trajectory,wang2021f,ma2019trafficpredict,chandra2019traphic,choi2021shared,shi2021sgcn,shafiee2021introvert,sun2020recursive,marchetti2020mantra,sun2020reciprocal,mohamed2020social}. They evaluate the models on benchmarks collected from real world using the average $\ell_2$ distance between ground truth and predicted trajectories as the key metric.
However, few studies evaluate trajectory prediction models from the perspective of security or analyze the robustness against adversarial examples.
For trajectory prediction, if the adversary can control the position of a vehicle close to the target AV, e.g., by driving the vehicle along a crafted trajectory, the adversary can influence the AV's trajectory prediction and driving behaviors.

To bridge this gap, we propose new white/black box adversarial attacks on trajectory prediction, which adds minor perturbation on normal trajectories to maximize the prediction error.
Compared with adversarial attacks on image/video classification, attacking trajectory prediction is unique in two aspects.
First, the attack requires naturalness~\cite{wang2021human} of the adversarial examples.
Adversarial trajectories are natural if they obey physical rules and are possible to happen in the real world.
With naturalness, the trajectories can be reproduced by the attacker-controlled vehicle in the real world and cannot be easily classified as anomaly by AVs.
% If the trajectories observed by AVs are natural in terms of physical dynamics and possible to happen in the real world, AVs regard them as normal trajectories.
% Since the dimension of trajectory data is low, it is inevitable that one can distinguish the normal trajectory and its perturbed version. However, indistinguishability is not necessary for stealthy attacks. 
% As long as the observed trajectories are natural in terms of physical dynamics and possible to happen in the real world, AVs regard them as normal trajectories. 
To realize naturalness, we enforce constraints on physical properties (e.g., velocity and acceleration) of the perturbed trajectory during optimization solving.
Second, we need to define optimization objectives that are semantically-attractive for attackers targeting trajectory prediction. To this end, we find multiple attractive attack dimensions can co-exist even for the same scenario (e.g., causing the predicted trajectory to deviate laterally or longitudinally are both of interest to attackers in AV context). Thus, in our attack design we consider different metrics of prediction error as optimization objectives, e.g., average lateral/longitudinal deviation to four different directions.

%we need to define  different optimization objectives 

%various different objectives corresponding to different attack goals.
% Second, the attack maximizes various metrics of prediction error to achieve different attack goals.
%To affect AV's behavior, the adversary intends to deviate the predicted trajectory to one direction according to the specific scenario.
%For instance, to fool the AV to think that the adversary vehicle is changing to the left lane, the predicted trajectory should have large deviation to lateral left direction.
%To realize the targeted attack, we maximize different metrics of prediction error, e.g., average lateral/longitudinal deviation to four different directions.

% First, the perturbation should lead to real-world impacts such as wrong decisions made by the AV. We optimize the perturbation towards various goals, e.g., maximizing the deviation towards lateral or longitudinal directions. By specifying the perturbation goals, the adversary can achieve targeted attacks. For instance, by maximizing the lateral deviation to the left, the perturbed trajectory spoof the fake information to the AV that a nearby vehicle is changing to the left lane. 
% Second, the perturbed trajectory must be possible to happen in the real world. We set hard constraints on the velocity and acceleration of the perturbed trajectory and enforce the hard constraints during the iterative optimization solving. In this way, the adversarial trajectories obey physical rules and can be reproduced by real cars.

We evaluate the proposed attacks on 10 different combinations of prediction models~\cite{fqa, grip++, trajectron++} and trajectory datasets~\cite{apolloscape, ngsim, nuscenes}.
The results show that the adversarial perturbation can substantially increase the prediction error by around 150\%. 62.2\% of attacks cause prediction to deviate by more than half of the lane width, which are likely to significantly alter AV's navigation decisions.
In addition, we thoroughly analyze how various factors impact the attack results and make recommendations for improvements of implementing trajectory prediction such as leveraging map information and driving rules.
% Our case studies present that such adversarial examples can cause real-world safety issues such as making the AV take a hard brake.
We also explore mitigation mechanisms against adversarial trajectories through data augmentation and trajectory smoothing, which reduce the prediction error under attacks by 28\%.
In general, our work exposes the necessity of evaluating the worst-case performance of trajectory prediction. The hard cases involving natural but adversarial trajectories generated by attacks have critical safety concerns (e.g., causing hard brakes or even collisions) as demonstrated by our case studies.

Our main contributions are summarized as follows:
\vspace{-0.5em}
\begin{itemize}[leftmargin=*] 
\setlength{\itemsep}{1pt}
\setlength{\parskip}{1.5pt}
\item We propose the \emph{first} adversarial attack and adversarial robustness analysis on trajectory prediction for AVs considering real-world constraints and impacts.

\item We report a thorough evaluation of adversarial attacks on various prediction models and trajectory datasets.

\item We explore mitigation methods against adversarial examples via data augmentation and trajectory smoothing.
\end{itemize}
\section{Background and Related Work}

\noindent
\textbf{Autonomous vehicles}.
The autonomous vehicle is a cyber-physical system where the perception module learns the surrounding environment through sensors, the planning module makes decisions on the driving behaviors, and the control module physically operates the vehicle. 
Within this pipeline, trajectory prediction is required by AV systems (e.g., Baidu Apollo~\cite{apollo}, Autoware~\cite{autoware}) as a part of the perception module. It predicts the future trajectories of nearby moving objects such as vehicles and pedestrians, which are important inputs for the planning module. 
% For instance, when AV is about to do an unprotected right turn at an intersection and it correctly predicts that a car from the opposite direction is driving through, the AV will yield the right-of-way according to the driving rules.
Hence, accurate trajectory prediction is critical for safe AV driving.

\noindent
\textbf{Trajectory prediction models}.
Trajectory prediction, for AVs especially, is the problem of predicting future spatial coordinates of various road agents such as vehicles and pedestrians. These models are usually deep neural networks accepting spatial coordinates of observable road agents in the past a few seconds as the primary input and may extend the input by leveraging auxiliary features (e.g., vehicles' heading)~\cite{grip++,choi2021shared}, interaction among road agents~\cite{fqa,chandra2020forecasting,yuan2021agentformer}, physical dynamics~\cite{trajectron++}, or semantic maps~\cite{trajectron++,phan2020covernet,deo2020trajectory,narayanan2021divide} to improve the accuracy of prediction.
The existing evaluation metrics include Average Displacement Error (ADE), Final Displacement Error (FDE)~\cite{chandra2019robusttp}, off-road rate~\cite{nuscenes}, etc. These metrics reflect the average performance of trajectory prediction in the testing datasets.
Instead, we focus on adversarial robustness and the worst-case performance of trajectory prediction algorithms.

\noindent
\textbf{Adversarial robustness}.
Since deep learning models are generally vulnerable to adversarial examples, various studies analyze the adversarial robustness of neural networks~\cite{carlini2019evaluating,croce2020reliable,dong2020benchmarking,sun2020adversarial,sun2021adversarially}. In AV systems, studies show that tasks such as object detection~\cite{cao2019adversarial,sun2020towards}, object tracking~\cite{jia2020fooling}, and lane detection~\cite{sato2021dirty} can be affected by perturbing sensor signals or adding physical patches. However, no existing work studied the adversarial robustness of trajectory prediction.

\noindent
\textbf{Testing of trajectory prediction}.
AdvSim~\cite{wang2021advsim} leverages adversarial machine learning (AML) to generate high-risk driving scenarios for end-to-end testing AV systems. Instead, we aim to understand the vulnerabilities in prediction algorithms specifically.
Saadatnejad et. al.~\cite{saadatnejad2021socially} analyzes the inaccuracy of attention mechanisms in trajectories prediction algorithms but does not consider safety impact on real-world applications. We are the first to bridge adversarial robustness and real-world systems like AVs.
\section{Problem Formulation}
\label{sec:design}

In this section, we first introduce the formulation of the trajectory prediction task (\S\ref{sec:notation}). We then propose the attack model (\S\ref{sec:attack-model}), metrics of attack impact (\S\ref{sec:metrics}).

\subsection{Trajectory Prediction}
\label{sec:notation}

In this work, we focus on trajectory prediction which is executed repeatedly at a fixed time interval and makes one prediction at each time frame according to the current/history state of all observable objects (i.e., vehicles/pedestrians).
First, we denote the state of an object $i$ at time $t$ as $s_t^i$, including information of spatial coordinates and other optional features.
The trajectory of object $i$ is a sequence of object states from time frame $t_1$ to $t_2$ (inclusively), denoted as $s_{t_1:t_2}^i=\{s_{t_1}^i, \dots, s_{t_2}^i\}$.
At each time frame, the prediction algorithm consumes the history trajectories of objects to predict their future trajectories, which are optimized towards having the same distribution as the ground-truth future trajectories.
At time frame $t$, we denote the number of observed objects as $N$ and denote the number of time frames in history and future trajectories as $L_I$ and $L_O$ respectively.
% Then history trajectories are $H_t=\{H^{1}_t, H^{2}_t, \dots, H^{N}_t\}$ ($H^{i}_t=s_{t-L_I+1:t}^{i}$); predicted trajectories are $P_t=\{P^{1}_t, P^{2}_t, \dots, P^{N}_t\}$; and future trajectories are $F_t=\{F^{1}_t, F^{2}_t, \dots, F^{N}_t\}$ ($F^{i}_t=s_{t+1:t+L_O}^{i}$).
Then history trajectories are $H_t=\{H^{i}_t=s_{t-L_I+1:t}^{i} | i \in [1, N]\}$; ground-truth future trajectories are $F_t=\{F^{i}_t=s_{t+1:t+L_O}^{i} | i \in [1, N]\}$; predicted trajectories are $P_t=\{P^{i}_t=p_{t+1:t+L_O}^{i} | i \in [1, N]\}$ ($p_t^i$ is the predicted state of object $i$ at time $t$).

\subsection{Attack Model}
\label{sec:attack-model}

\begin{figure}[!t]
\centering
\includegraphics[width=0.40\textwidth, height=10em]{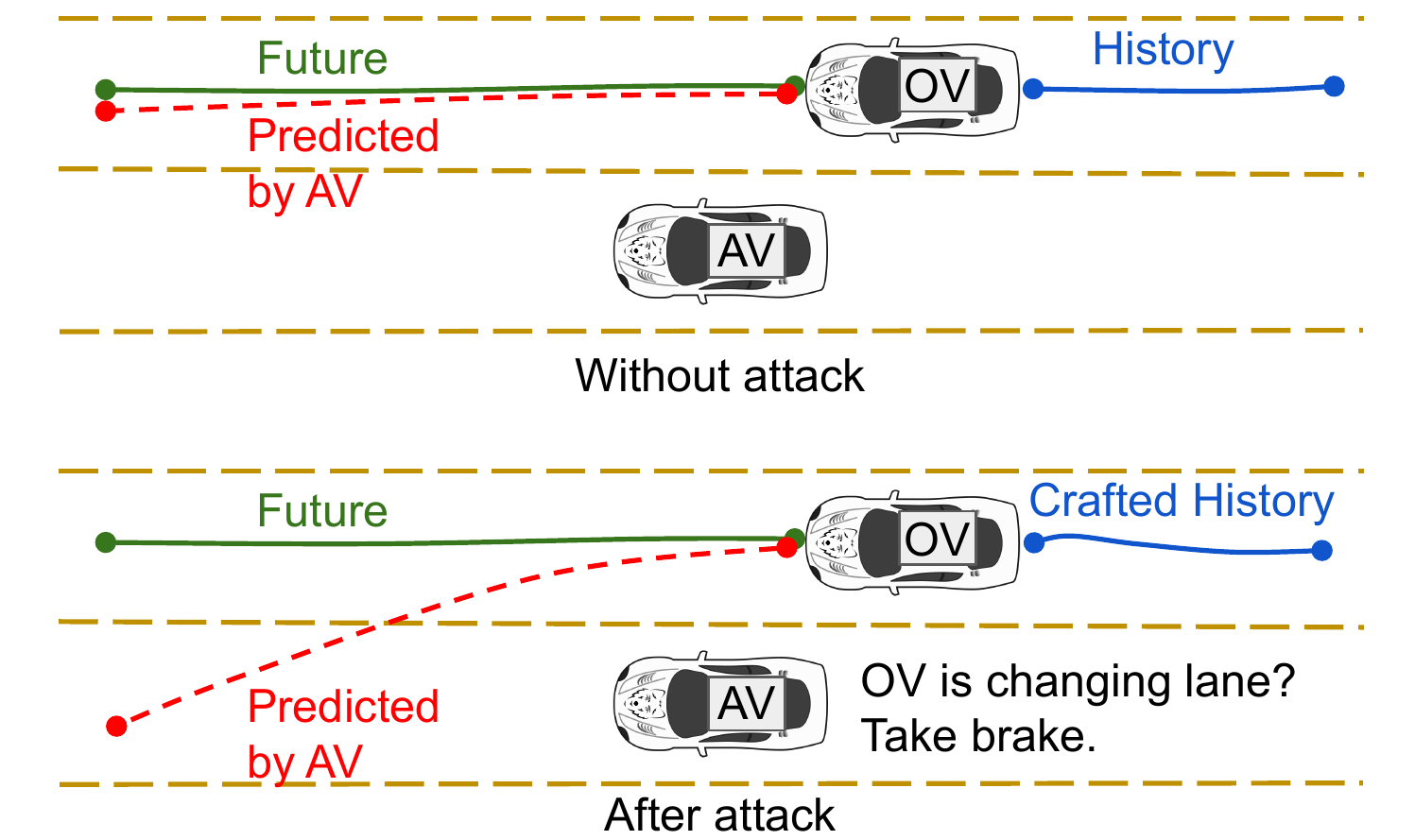}
\caption{An example of attack scenarios on trajectory prediction: spoofing lane-changing behavior.}
\label{fig:attack-model}
\end{figure}

% In order to generate adversarial examples that compromise prediction accuracy, the adversary must influence the input of trajectory prediction, which is fundamentally the trajectories (i.e., sequences of spatial locations) observed by the AV.
In this paper, we focus on the setting where the adversary drives one vehicle called ``the other vehicle'' (OV) along a crafted trajectory. The AV observes the OV and applies iterative trajectory prediction which produces the predicted trajectory of the OV at each time frame.
The adversary controls the OV's whole trajectory to maximize the prediction error or make the AV take unsafe driving behaviors.
Figure~\ref{fig:attack-model} demonstrates one example of the attack. By driving along a crafted trajectory, the OV seems be changing its lane in the AV's prediction while the OV is actually driving straightly. Given the high-error prediction, the AV takes brake to yield the OV. If the AV brakes on the highway, it is a serious safety hazard that may cause rear-end collisions. 

In a realistic attack scenario, the adversary needs to access all parameters or only APIs of the prediction model equipped by the victim AV for white-box and black-box settings respectively. 
% \st{may need to justify the feasibility of such access permission. e.g., open-source models/AV systems} 
When the adversary drives the OV close to the AV and is ready to attack, he/she first selects a future period and predicts the trajectories of surrounding on-road objects in that period, which are necessary inputs for generating the adversarial trajectory. The adversary then computes the adversarial trajectory for that future period and drives exactly on it (e.g., the adversary can control the OV using software).
Though the generation of adversarial trajectories relies on predicted future knowledge that is not guaranteed to be accurate, the attack is still effective since the attack impact is mostly determined by the OV's trajectory itself. The experimental evidences are in the appendix.

In addition, adversarial trajectories must satisfy the following requirements to ensure naturalness. 
First, the trajectory obeys the physical rules. The physical properties (e.g., velocity, acceleration) must be bounded so that real vehicles can reproduce the trajectory.
Second, the trajectory represents normal driving behavior instead of ruthless driving.

\subsection{Evaluation Metrics} 
\label{sec:metrics}

We use six metrics to evaluate the prediction error. Also, an attack is effective if the prediction error is significantly raised after the adversarial perturbation. We first include two metrics that are commonly used in related works~\cite{chandra2019robusttp}. 
(1) Average displacement error (\textbf{ADE}). The average of the root mean squared error (RMSE) between the predicted and ground-truth trajectory. 
(2) Final displacement error (\textbf{FDE}). The RMSE between the predicted and ground-truth trajectory position at the last predicted time frame.

However, the above two metrics are not sufficient to evaluate the impact of targeted attacks.
For instance, to spoof lane-changing behavior to left (Figure~\ref{fig:attack-model}), the predicted trajectory should have a deviation to left.
Similarly, to spoof a fake acceleration, the deviation should be towards the front of the longitudinal direction.
For the above attacks, only the deviation to a specific direction is counted as an effective attack impact.
Therefore, we design four extra metrics including
the average deviation towards \textbf{left}/\textbf{right} side of lateral direction and \textbf{front}/\textbf{rear} side of longitudinal direction. The metrics are formally defined as:

\vspace{-1em}
\begin{equation}
\label{eq:objective}
D(t,n,R) = \frac{1}{L_O} \sum_{\alpha=t+1}^{t+L_O} (p_\alpha^n - s_\alpha^n)^T \cdot R(s_{\alpha+1}^n, s_\alpha^n),
\end{equation}
\vspace{-1em}

where $t$ denotes the time frame ID, $n$ denotes the target vehicle ID, $p$ and $s$ are binary vectors representing predicted and ground-truth vehicle locations respectively, and $R$ is a function generating the unit vector to a specific direction. We approximate longitudinal direction as $s_{\alpha+1}^n - s_\alpha^n$.

According to our case studies, we believe that half of the lane width (e.g., about 1.85 meters in datasets we use) is a threshold of deviation to cause real-world impacts in a high probability.
If the average deviation exceeds the threshold, the last predicted trajectory location is likely on a different lane, which may lead to different decisions of the AV.
% For convenience, we name the six metrics \textbf{ADE}, \textbf{FDE}, \textbf{left}, \textbf{right}, \textbf{front}, \textbf{rear}, respectively.

\section{Adversarial Example Generation}
\label{sec:attack}

We design white box and black box attacks against trajectory prediction following definitions in \S\ref{sec:notation}.

\begin{figure}[!t]
\centering
\includegraphics[width=0.45\textwidth]{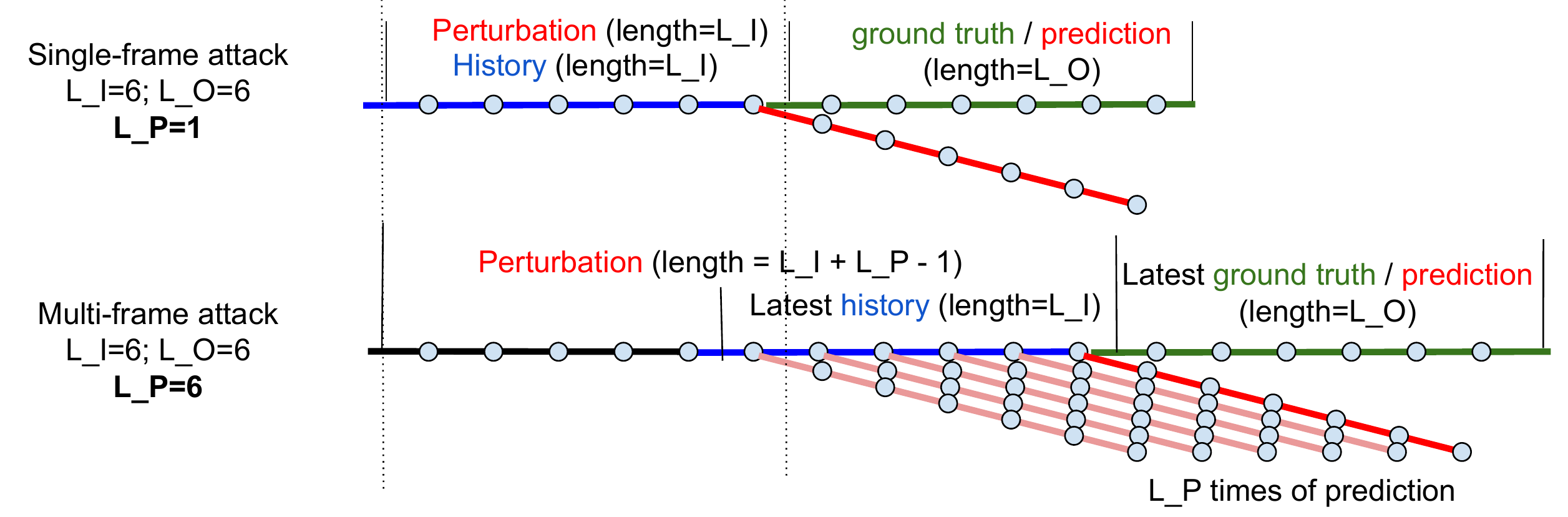}
\caption{Illustration of single/multi-frame attack. The dots are observed/predicted trajectory locations in different time frames.}
\label{fig:parameter_demo}
\end{figure}

\noindent
\textbf{Perturbation.}
We generate the adversarial trajectory by adding minor perturbation on normal trajectories.
The perturbation changes the spatial coordinates (x-y location) and features that are calculated from locations (e.g., velocity) correspondingly.
As shown in Figure~\ref{fig:attack-model}, perturbation is applied on the history trajectory in length of $L_O$ to control the prediction. Since only the prediction of the current time frame is considered, we name this attack \textbf{single-frame} attack.
However, real-world safety issues usually happen in a longer time sequence. Therefore, we generalize the attack to \textbf{multi-frame}, as shown in Figure~\ref{fig:parameter_demo}.
We define parameter $L_P$ to represent the number of predictions considered in the attack objective. 
Given $L_P$, one attack scenario include $L_I+L_O+L_P-1$ time frames. 
We apply perturbation on the first $L_I + L_P - 1$ time frames so that predicted trajectories on time frames $\{L_I, L_I+1, \dots, L_I+L_P-1\}$ ($L_P$ frames in total) are controlled by adversarial trajectories.
We maximize the total prediction error in these $L_P$ time frames to launch the multi-frame attack.
% The detailed method of generating the perturbation is in \S\ref{sec:attack}.

% Given the normal trajectories and parameters, the adversarial attack aims to optimize the perturbation to maximize the prediction error (six metrics in \S\ref{sec:metrics}) on the OV.
% For instance, the perturbation in Figure~\ref{fig:attack-model} is optimized to maximize lateral deviation to left. 

\noindent
\textbf{Objective}. The optimization has six different objective functions corresponding to the six metrics of prediction error. The objective is the negation of the average prediction error in all considered time frames:

\vspace{-0.8em}
\begin{equation}
\label{eq:objective}
L(n,f)=-\frac{1}{L_P} \sum_{\alpha=L_I}^{L_I+L_P-1} f(P_{\alpha}^{n}, F_{\alpha}^{n}),
\end{equation}
\vspace{-0.8em}

where $f$ denotes the function calculating one of the six metrics; $n$ is ID of the target vehicle (i.e., OV); $P$ and $F$ represents predicted and future trajectories (\S\ref{sec:notation}).

\noindent
\textbf{Hard constraints of perturbation.} As mentioned in \S\ref{sec:attack-model}, the perturbed trajectories must be physically feasible and not perform dangerous driving behaviors.
To enforce this requirement, we design constraints that need to be satisfied by any perturbation.

First, the bounds of physical properties.
We traverse all trajectories in the testing dataset to calculate the mean ($\mu$) and standard deviation ($\sigma$) of (1) scalar velocity, (2) longitudinal/lateral acceleration, and (3) derivative of longitudinal/lateral acceleration.
For each $\mu$ and $\sigma$, we check that the value of perturbed trajectories does not exceed $\mu \pm 3\sigma$. Assuming the physical properties are in the normal distribution, this range covers 99.9\% of the dataset but excludes outliners.
We also manually check that the bounds are physically reasonable. 
% Also, physical constraints are not the only restriction we considered (\S{4}). We have hard bounds on the deviation of each trajectory point to further ensure that the adversarial trajectory does not go to another lane.
In addition, when checking the physical constraints, three ground-truth trajectory points are involved before and after the perturbed part thus the boundary of normal and perturbed trajectories is natural.
% In this way, we ensure their physical properties satisfy the real-world physical rules that the dataset implies.

Second, the bound of deviation on each trajectory location. We set the maximum deviation as 1 meter by default. Given that the urban lane width is around 3.7 meters in datasets we use and the width of cars is about 1.7 meters, the 1-meter deviation is an upper bound for a car not shifting to another lane if it is normally driving in the center of the lane. The bound (1) rules out ruthless driving by preserving the original driving behavior and (2) is a parameter for tuning the stealth level of the attack.

We enforce the constraints by reducing the perturbation whenever the constraints are violated. Given the perturbation $\Delta$, the history trajectory of the target vehicle $H^{n}$, and the constraint function $C$, we calculate the maximum coefficient $0 \le \theta \le 1$ which reduces the perturbation to $\theta \Delta$ while $\theta \Delta$ satisfies all constraints. Formally speaking, the calculation of $\theta$ is an optimization problem:

\vspace{-0.8em}
\begin{equation}
\label{eq:constraint}
\begin{aligned}
\max & \ \theta \\
\textrm{s.t.} & \ C(H^{n}+\theta \Delta) \land 0 \le \theta \le 1
\end{aligned}
\end{equation}
\vspace{-0.8em}

% \begin{algorithm}[t]
% \small
% \SetAlgoLined
% \KwIn{Hyper parameters $L_I$, $L_O$, $L_P$, iteration number $K$, learning rate $l_r$; vehicle trajectories $s_{1:L_I+L_O+L_P-1}^{i}$ ($1 \le i \le N$); tested model $M$, loss function $f$; constraint $C$; target vehicle id $1 \le n \le N$.}
% \KwOut{The best perturbation $\Delta_{best}$; the best loss $L_{best}$.}
% Randomly initialize $\Delta^{(1)}$; $L_{best} \gets +\infty$\;
% \For{$k=1, 2, \dots, K$} {
%   Perturb $s$ to $\delta(s)$: $\delta(s)_{1:L_I+L_P-1}^{n} = s_{1:L_I+L_P-1}^{n} + \Delta^{(k)}$\; 
%   \For{$\alpha=1, 2, \dots, L_P$} {
%     $H_{\alpha} \gets \{\delta(s)_{\alpha:\alpha+L_I-1}^{i} | i\in 1\dots N\}$\;
%     $F_{\alpha} \gets \{\delta(s)_{\alpha+L_I:\alpha+L_I+L_O-1}^{i} | i\in 1\dots N \}$\;
%     $P_{{\alpha}} \gets M(H_{{\alpha}})$\;
%   }
%   $L^{(k)} \gets \frac{1}{L_P} \sum_{\alpha=1}^{L_P} f(P_{\alpha}, F_{\alpha})$\;
%   \If{$L^{(k)} < L_{best}$} {
%     $L_{best} \gets L^{(k)}$\;
%     $\Delta_{best} \gets \Delta^{(k)}$\;
%   }
%   $\Delta^{(k+1)} \gets \theta (\Delta^{(k)} + l_r \nabla L$) ($\theta$ is solved by $\max \theta\ \textrm{s.t.}\ C(H_t+\Delta^{(k+1)}) \land 0 \le \theta \le 1$)\;
% }
% \caption{Generation of optimized perturbation.}
% \label{alg:optimize}
% \end{algorithm}

\noindent
\textbf{White box optimization}.
We design our white box optimization method based on Projected Gradient Descent (PGD)~\cite{madry2017towards}. The process can be summarized as follows.

The perturbation is initialized randomly.
In each iteration, we first enforce the hard constraint on the current perturbation following Equation~\ref{eq:constraint} and then add the perturbation on the original history trajectories of the target vehicle $n$ which is observed by the AV. Then $L_P$ times of prediction is executed on the perturbed trajectory data and the loss of the iteration is calculated using Equation~\ref{eq:objective}.
Next, the algorithm updates the perturbation using gradient descents.
Finally, the algorithm produces the best perturbation which can transform the original scenario to the worst-case scenario with the maximized prediction error.

\noindent
\textbf{Black box optimization}.
Methods based on gradient descent are not always feasible because the trajectory prediction models may have non-differentiable layers. Hence, we design a black box attack method based on Particle Swarm Optimization (PSO)~\cite{kennedy1995particle} which requires only the API of model inference instead of gradients.
PSO is an optimization method by iteratively improving solution candidates (i.e., particles) with regards to a given measure of quality in the search space. In this case, each particle is one candidate of perturbation, the measure of quality is defined by objective function (Equation~\ref{eq:objective}), and the search space is defined by the hard constraints (Equation~\ref{eq:constraint}).
% Since the perturbation has a low dimension, PSO is also effective for finding optimal solutions. 

\section{Mitigation Mechanisms}
\label{sec:defense}

From our observation of attack results (\S\ref{sec:experiments}), the adversarial trajectories frequently change acceleration, which is a rare pattern in normal trajectories.
Based on this phenomenon, we design mitigation methods as follows.

\noindent
\textbf{Data augmentation}. Since normal trajectories in the training dataset are mostly smooth with stable acceleration, adversarial trajectories have a different data distribution. Hence, we apply data augmentation to inject adversarial patterns in the training data.
During the training, we add random perturbation on randomly selected trajectories while the perturbation satisfies the hard constraints defined in \S\ref{sec:attack}.
We do not adopt adversarial training because of its limitations such as the high cost of training and poor generality in terms of attack goals.

\noindent
\textbf{Train-time trajectory smoothing}. Since the unstable velocity or acceleration is a key pattern of adversarial trajectories, we can partially remove the adversarial effect by smoothing the trajectories. We apply trajectory smoothing on both training and testing data. There are various choices of smoothing algorithms and we use a simple linear smoother based on convolution in our experiments.
This mitigation relies on the physical properties of trajectories instead of gradient obfuscation~\cite{athalye2018obfuscated}. Hence, it does not matter if the attacker knows the gradients of the smoothing.

\noindent
\textbf{Test-time detection and trajectory smoothing}. The above two mitigation methods modify the distribution of training data so that one needs to retrain the model. To make the mitigation easier to deploy, we propose another method that only smooths trajectories during inference time if the trajectory is detected as adversarial. 
% However, it is a problem that the test-time trajectory smoothing alters the data distribution of testing data to be different from the training data thus prediction performance on normal trajectories is damaged. To alleviate the drawback, we add an attack detection module and only apply trajectory smoothing on detected adversarial trajectories.
We design two methods for detecting adversarial trajectories.
First, SVM classifier~\cite{steinwart2008support}. We extract the magnitude and direction of the acceleration as features to fit an SVM model for classifying normal and adversarial trajectories. Second, rule-based detector. We calculate the variance of acceleration over time frames and if the variance is higher than a threshold the trajectory is detected as adversarial.
% The smoothing algorithm is the same as the train-time smoothing method.

\section{Experiments}
\label{sec:experiments}

\begin{table}[t]
\scriptsize
\caption{Summary of datasets.}
  \label{tab:datasets}
  \vspace{-0.5\baselineskip}
  \centering
  \begin{tabular}{| c | c | c | c | c | c |}
    \noalign{\global\arrayrulewidth1pt}\hline\noalign{\global\arrayrulewidth0.4pt}
    Name & Scenario & Map & $L_I$ & $L_O$ & Freq. (Hz)\\
    \noalign{\global\arrayrulewidth1pt}\hline\noalign{\global\arrayrulewidth0.4pt}
    Apolloscape~\cite{apolloscape} & Urban & $\times$ & 6 & 6 & 2 \\
    \hline
    NGSIM~\cite{ngsim} & Highway & $\times$ & 15 & 25 & 5 \\
    \hline
    nuScenes~\cite{nuscenes} & Urban & $\surd$ & 4 & 12 & 2 \\
    \noalign{\global\arrayrulewidth1pt}\hline\noalign{\global\arrayrulewidth0.4pt}
  \end{tabular}
\end{table}

\begin{table}[t]
\scriptsize
\caption{Summary of models.}
  \label{tab:models}
  \vspace{-0.5\baselineskip}
  \centering
  \begin{tabular}{| >{\centering}m{5.5em} | >{\centering}m{8em} | c | >{\centering\arraybackslash}m{7em} |}
    \noalign{\global\arrayrulewidth1pt}\hline\noalign{\global\arrayrulewidth0.4pt}
    Name & Input features & Output format & Network \\
    \noalign{\global\arrayrulewidth1pt}\hline\noalign{\global\arrayrulewidth0.4pt}
    FQA~\cite{fqa} & location & single-prediction & LSTM \\
    \hline
    GRIP++~\cite{grip++} & location + heading & single-prediction & Conv + GRU \\
    \hline
    Trajectron++ \cite{trajectron++} & location + physical dynamics + map & multi-prediction & Conv + LSTM + GRU + GMM \\
    \noalign{\global\arrayrulewidth1pt}\hline\noalign{\global\arrayrulewidth0.4pt}
  \end{tabular}
\end{table}
\begin{table*}[t]
\scriptsize
\caption{Average prediction error before and after single-frame white box attack.}
  \label{tab:loss}
  \vspace{-0.5\baselineskip}
  \centering
\begin{tabular}{|c|c|c|c|c|c|c|c|}
\noalign{\global\arrayrulewidth1pt}\hline\noalign{\global\arrayrulewidth0.4pt}
\multirow{2}{*}{Model}     & \multirow{2}{*}{Dataset}   & \multicolumn{1}{c|}{ADE (m)}  & \multicolumn{1}{c|}{FDE (m)}  & \multicolumn{1}{c|}{Left (m)}  & \multicolumn{1}{c|}{Right (m)} & \multicolumn{1}{c|}{Front (m)} & \multicolumn{1}{c|}{Rear (m)}   \\
\cline{3-8}
      &     & Normal / Attack  & Normal / Attack  & Normal / Attack  & Normal / Attack  & Normal / Attack  & Normal / Attack  \\
\noalign{\global\arrayrulewidth1pt}\hline\noalign{\global\arrayrulewidth0.4pt}
GRIP      & Apolloscape & 1.97 / 7.14 & 3.18 / 10.74  & -0.0128 / 2.65 & 0.0128 / 2.38 & -0.0154 / 4.80 & 0.0154 / 5.49 \\
\hline
GRIP      & NGSIM   & 7.29 / 9.24 & 12.8 / 16.5  & -0.175 / 1.10 & 0.175 / 0.311 & -5.30 / -3.76  & 5.30 / 5.85 \\
\hline
GRIP      & nuScenes  & 5.46 / 8.32 & 10.3 / 15.4  & 0.233 / 1.53 & -0.233 / 1.44  & -1.04 / 1.50 & 1.04 / 3.76 \\
\hline
FQA     & Apolloscape & 2.37 / 5.64 & 3.82 / 8.78 & 0.0479 / 2.10 & -0.0479 / 1.86 & -0.387 / 3.11 & 0.387 / 3.84 \\
\hline
FQA     & NGSIM   & 5.44 / 7.05 & 9.45 / 12.2  & 0.205 / 1.19 & -0.205 / 0.787 & -1.11 / 2.12 & 1.11 / 2.16 \\
\hline
FQA     & nuScenes  & 5.28 / 7.83  & 10.0 / 14.4 & 0.229 / 1.18 & -0.229 / 0.736 & -0.814 / 1.79 & 0.814 / 3.347 \\
\hline
Trajectron++  & Apolloscape & 1.31 / 7.11 & 2.28 / 10.8  & 0.00679 / 4.14 & -0.00679 / 3.69 & -0.183 / 4.56 & 0.183 / 5.24 \\
\hline
Trajectron++  & NGSIM   & 2.31 / 9.64  & 4.49 / 16.9  & -0.219 / 4.39 & 0.219 / 4.62 & -0.0855 / 4.38 & 0.0855 / 4.28 \\
\hline
Trajectron++  & nuScenes  & 5.31 / 8.73 & 11.2 / 17.1  & -0.331 / 2.10 & 0.331 / 2.68 & -1.57 / 2.25 & 1.57 / 5.01 \\
\hline
Trajectron++(map) & nuScenes  & 2.69 / 6.81 & 5.39 / 11.8  & -0.107 / 2.118 & 0.107 / 2.25 & -0.526 / 3.79  & 0.526 / 4.48 \\
\noalign{\global\arrayrulewidth1pt}\hline\noalign{\global\arrayrulewidth0.4pt}
\end{tabular}
\end{table*}

In this section, we evaluate proposed attack/mitigation methods and analyze the results.

\subsection{Experimental Setting}
\label{sec:eval-setup}

\noindent
\textbf{Datasets}. We summarize the characteristics of the three datasets we use in Table~\ref{tab:datasets}. We select history trajectory length ($L_I$) and future trajectory length ($L_O$) following the recommendation from the dataset's authors.
% For instance, when testing prediction on \emph{Apolloscape}, we let prediction models observe 6 time steps and predict the future 6 steps. Since the interval between steps is 0.5 seconds in \emph{Apolloscape}, it means predicting 3-second trajectories future by observing 3-second trajectories.
% Note that only \emph{nuScenes} provide semantic maps.
We randomly select 100 scenarios as test cases from each dataset.

\noindent
\textbf{Models}. We summarize three prediction models we use in Table~\ref{tab:models}, which are open-source state-of-the-art models at the time of the experiments.
% In terms of input features, \emph{FQA} only uses locations of vehicles (x-y coordinates) as model input while \emph{Trajectron++} leverages the most features including velocity, heading, map information, etc.
% In terms of the model output, both \emph{FQA} and \emph{GRIP++} produce one single future trajectory for each vehicle (i.e., single-prediction) while \emph{Trajectron++} produces multiple trajectories with probabilities (i.e., multi-prediction). 
Note that \emph{Trajectron++} model produces multiple predicted trajectories with probabilities (i.e., multi-prediction) and we select the trajectory with the highest probability as the final result.
Also, \emph{Trajectron++} requires the semantic map as input which is only available in \emph{nuScenes} dataset. Therefore, we prepare two versions of \emph{Trajectron++}. \emph{Trajectron++} (w/o map) is evaluated on all datasets while \emph{Trajectron++} (w/ map) is evaluated on \emph{nuScenes}. For each combination of models and datasets, we train the model using fine-tuned hyperparameters.
% All experiments are done on a server with four Nvidia RTX 2080Ti GPUs and Intel Xeon Silver 4110 2.10GHz CPUs.

\noindent
\textbf{Implementation details}. Our implementation is open sourced at \url{https://github.com/zqzqz/AdvTrajectoryPrediction}. For the PGD-based white box attack, we use Adam optimizer with a learning rate of 0.01 and set the maximum iteration to 100. For PSO-based black box attack, we set the number of particles to 10, inertia weight to 1.0, acceleration coefficients to (0.5, 0.3), and the maximum iteration to 100.
For mitigation, we use convolution-based linear smoother for trajectory smoothing and the SVM implementation in \emph{scikit-learn}~\cite{pedregosa2011scikit} for anomaly detection. More details are in the appendix.
% For trajectory smoothing, we use a simple convolution kernel $(\frac{1}{3},\frac{1}{3},\frac{1}{3})$, which takes the mean of three trajectory positions as the smoothed position at the middle time frame.
% For detection, the SVM model uses RBF kernel.
% The thresholds for detection are selected to achieve the best Area Under Curve (AUC) score of Receiver Operating Characteristics (ROC) curve on a set of data other than the test cases, which includes normal and anomaly trajectories (generated by random perturbation).  

\subsection{Attack Results}
\label{sec:eval-attack}

% \begin{figure*}[!t]
% \centering
% \includegraphics[width=18cm]{figures/average_loss.pdf}
% \caption{Prediction error (in meters) in all experiments. Each row refers to one model while each column refers to one dataset. In each sub-figure, we draw the 6 metrics under both single-frame and multi-frame mode, before and after the perturbation.}
% \label{fig:average-loss}
% \end{figure*}

\begin{figure*}[t]
\centering
\includegraphics[width=0.75\textwidth]{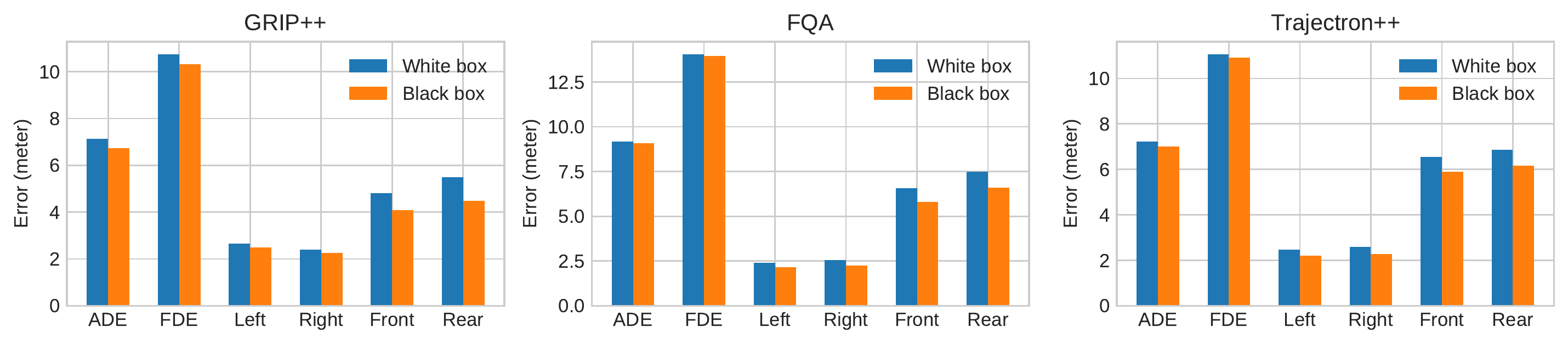}
\caption{White box v.s. block box attack (single-frame) on \emph{Apolloscape} dataset.}
\label{fig:blackbox}
\end{figure*}

\begin{figure*}[t]
\centering
\includegraphics[width=0.75\textwidth]{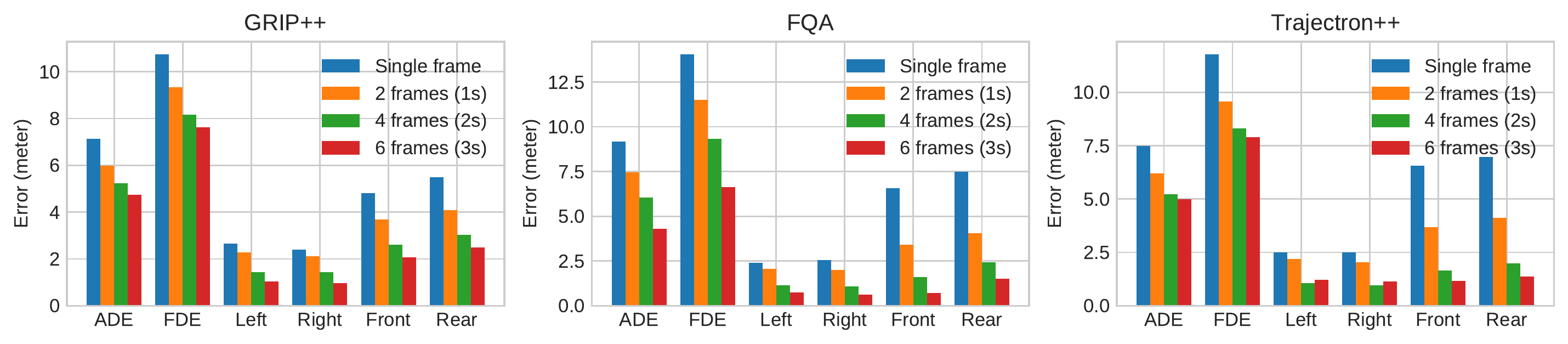}
\caption{Attack results (white-box, single-frame, on \emph{Apolloscape}) w.r.t. different length of adversarial perturbation.}
\label{fig:attack_length}
\end{figure*}

\begin{figure*}[t]
\centering
\includegraphics[width=0.75\textwidth]{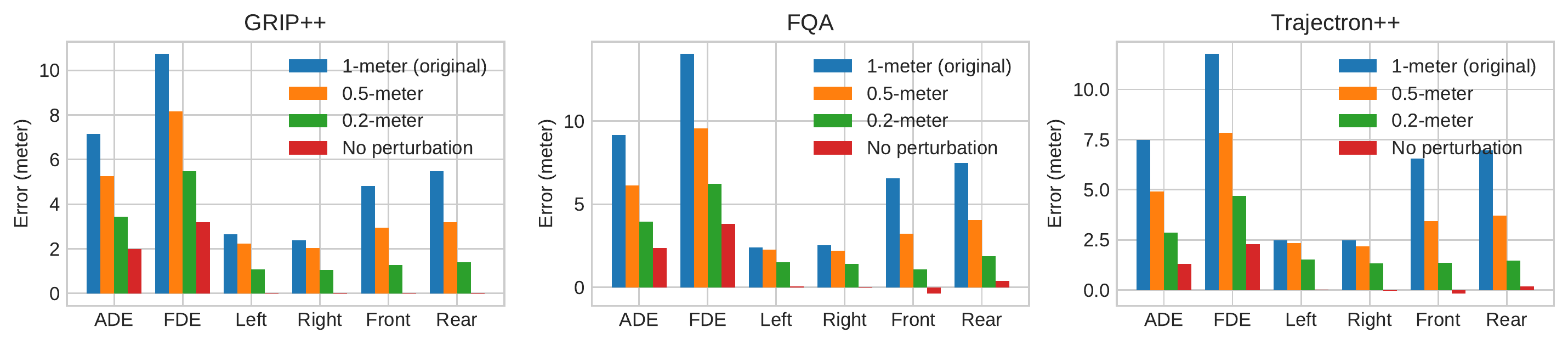}
\caption{Attack results (white-box, single-frame, on \emph{Apolloscape}) w.r.t. different bounds of largest deviation of trajectory positions.}
\label{fig:threshold}
\end{figure*}

\noindent
\textbf{General results}. For each model \& dataset combination, each test case, and each perturbation objective, we first execute the white-box attack in the setting of single-frame prediction ($L_P=1$). We assume the attacker acknowledges history trajectories of all objects on-road and discuss a more realistic setting in the appendix.
We present the average prediction error before and after perturbation in Table~\ref{tab:loss}.
Generally, the white-box adversarial perturbation is effective on all models and datasets. On average, ADE/FDE is increased by 167\%/150\%. The lateral/longitude deviation reaches 2.03/3.84 meters and 62.2\% of attacks are likely to cause real-world impact since their average deviations are larger than half of the lane width (i.e., 1.85 meters). 

\noindent
\textbf{Naturalness}. The adversarial trajectories can naturally happen in the real world in a reasonable probability.
One characteristic of adversarial trajectories is that they often change vehicle velocity or acceleration. However, a small fraction of normal trajectories share the same pattern and are indistinguishable from adversarial trajectories. 
The hard constraint (\S\ref{sec:attack}) also ensures that adversarial trajectories are physically feasible to reproduce by driving a real car.
Hence, the adversarial attack can be regarded as a method to discover the worst but realistic cases of prediction. 

Next, we will analyze factors that affect adversarial robustness in \S\ref{sec:eval-scenarios}, \S\ref{sec:eval-models}, and \S\ref{sec:eval-attack-methods}. 
The analysis uses experiments on dataset \emph{Apolloscape} as supporting evidence. As a baseline, in the single-frame white-box attack on \emph{Apolloscape}, ADE/FDE is increased by 239\%/206\% while lateral/longitude deviation reaches 2.49/6.31 meters on average (53.6\% of deviations are larger than 1.85 meters).
If not specified, the percentage rise/drop of prediction error is the average of six metrics.
% We also show transferability analysis in \S\ref{sec:eval-transfer} and the attack demonstration in a more realistic setting in \S\ref{sec:eval-realistic}.

\vspace{-1em}
\subsubsection{Different Scenarios}
\label{sec:eval-scenarios}
\vspace{-0.5em}

\noindent
\textbf{High-acceleration scenarios}. The prediction error is generally high in scenarios where vehicles have high acceleration. Taking metric ADE as an example, the prediction error in high-acceleration scenarios (average acceleration $> 1 m/s^2$) w/o and w/ perturbation are 31\% and 20\% higher than low-acceleration scenarios respectively.
Typical high-acceleration scenarios include turning at intersections or stopping at stop lines (details in the appendix).
All three models cannot foresee the driver's behavior in such scenarios thereby have extremely high prediction errors.
% Figure~\ref{fig:hard-scenarios} presents two examples of the hard scenarios where all three models have high prediction error.
% We present examples of such hard scenarios in the appendix~\ref{sec:hard-scenarios}.
The models need specific knowledge about traffic rules to make the model robust in those hard cases.

% \begin{figure*}[t]
% \centering
% \includegraphics[width=0.80\textwidth]{figures/density.pdf}
% \caption{Attack results (white-box, single-frame, on \emph{Apolloscape}) on scenarios with different density of traffic.}
% \label{fig:density}
% \end{figure*}

\noindent
\textbf{Density of traffic}. 
Prediction models commonly encode the interaction among objects as a graph representation to improve prediction accuracy. 
To study the factor of traffic density, we repeat the white box attack on three models using \emph{Apolloscape} dataset but drops 50\% or 100\% of objects other than the target vehicle. 
% \jc{no need to highlight the normal drop in texts.} 
% In normal cases, sparse traffic causes slightly worse performance (8\% larger error if 100\% other objects are dropped). 
% It shows that interaction representation improves prediction accuracy in normal cases.
However, the prediction error under the white-box attack appears independent of traffic density. The result shows that the worst-case prediction error under attacks is mostly determined by the target vehicle's trajectory rather than the surrounding objects. 
% The attacker controlling the target vehicle can potentially generate effective adversarial trajectories without knowledge of other vehicles.
% \textbf{Urban v.s. freeway}.

\vspace{-1em}
\subsubsection{Different Models}
\label{sec:eval-models}
\vspace{-0.5em}

\noindent
\textbf{Input representation}. Features independent of trajectory locations (e.g., maps) help to improve adversarial robustness. Comparing \emph{Trajectron++} w/ map against w/o map, ADE on original and perturbed trajectories are reduced by 22\% and 31\% respectively. By adding map information as input features, the weight on perturbed features is lower so that the prediction result is less sensitive to the perturbation.

Also, good prediction accuracy on normal trajectories does not necessarily lead to good adversarial robustness. Without perturbation, \emph{Trajectron++} has better prediction accuracy on all three datasets, thanks to its comprehensive input representation. 
With adversarial perturbation, however, \emph{Trajectron++} does not have the best accuracy. On dataset \emph{Apolloscape} and \emph{NGSIM}, \emph{FQA} has a better prediction error against perturbation. \emph{FQA}'s single LSTM makes predictions mainly based on the last two trajectory positions so that the perturbation on other parts of trajectories is not effective. Though \emph{Trajectron++} integrates rich features such as dynamics, the features are affected by perturbation on any trajectory position, which is a wide attack surface.

\vspace{-1em}
\subsubsection{Different Attack Methods}
\label{sec:eval-attack-methods}
\vspace{-0.5em}

\noindent
\textbf{White box v.s. black box}.
We conduct black box attacks on the three models and evaluate the attack result on the \emph{Apolloscape} dataset. We visualize the six metrics of prediction error in Figure~\ref{fig:blackbox}.
In general, black box attacks and white box attacks have a very similar performance.
Since the search space of the optimal perturbation is in a two-dimensional space (i.e., the spatial locations) and restricted by hard constraints (\S\ref{sec:attack}), the attacker can efficiently solve the optimization problem without knowledge of the model. 
White-box and black-box adversarial trajectories both have high variance of acceleration but perturbation of white-box attack is overall smaller because the guidance of gradient avoids part of unnecessary perturbation.

\noindent
\textbf{Length of adversarial perturbation}
As mentioned in \S\ref{sec:attack}, the attacker can launch attacks on continuous time frames by tuning the parameter $L_P$.
Besides the attack on single-frame attack ($L_P=1$) presented in Table~\ref{tab:loss}, we increase $L_P$ to the number of frames in up to 3 seconds as multi-frame attacks.
For instance, for \emph{Apolloscape} dataset which has a sampling frequency of 2 Hz, $L_P=6$ means all predictions in 3 seconds are controlled by adversarial trajectories.
We show white-box attack results with various $L_P$ of three models on \emph{Apolloscape} in Figure~\ref{fig:attack_length}.

Without perturbation, the single-frame and multi-frame prediction have similar performance because multi-frame prediction is a sequence of independent single-frame prediction.
However, under adversarial attacks, the average prediction error decreases as the length of adversarial perturbation increases.
The attack gets harder if the attacker intends to maintain the attack impact in a long continuous time period.
This is because of the extra constraint in multi-frame perturbation -- adversarial history trajectories in different time frames depends on the same perturbation vector thereby are not optimized separately.
For the 3-second attack, ADE/FDE is increased by 142\%/127\% and lateral/longitudinal deviation is 0.95/1.55 meters on average (22\% of deviations are larger than 1.85 meters).
% \jc{provide some conclusion and suggestions here. e.g., our results shed light on a promising design choice, which is to relatively increase the historic time series ....}

\noindent
\textbf{Bounds on the perturbation}. As mentioned in \S\ref{sec:attack}, the attacker can tune the deviation bound to generate attacks with different stealth levels. We launch white-box attacks under various bounds and the attack impact reduces as the bound decreases (as shown in Figure~\ref{fig:threshold}). However, the attack still effectively cause high prediction error using minor perturbations. When the deviations of trajectory locations are at most 0.2-meter, ADE/FDE is increased by 86\%/80\% and lateral/longitude deviation is 1.31/1.40 meters on average (27\% of deviations are larger than 1.85 meters).

\begin{table}[t]
\scriptsize
\caption{Attack transferability (white box, single-frame).}
  \label{tab:transfer}
  \vspace{-0.5\baselineskip}
  \centering
\begin{tabular}{|c|c|c|c|}
\noalign{\global\arrayrulewidth1pt}\hline\noalign{\global\arrayrulewidth0.4pt}
\backslashbox{Source}{Target} & FQA & GRIP++ & Trajectron++ \\
\hline
FQA & 100\% & 42.8\% & 16.5\% \\
\hline
GRIP++ & 60.8\% & 100\% & 15.5\% \\
\hline
Trajectron++ & 64.7\% & 39.5\% & 100\% \\
\noalign{\global\arrayrulewidth1pt}\hline\noalign{\global\arrayrulewidth0.4pt}
\end{tabular}
\end{table}

\vspace{-1em}
\subsubsection{Transferability}
\label{sec:eval-transfer}
\vspace{-0.5em}

We evaluate the transferability across the three models using \emph{Apolloscape} dataset. Since the attack impact is evaluated using quantified prediction errors instead of the binary judgment of attack success, we define a percentage score to measure the transferability.
When we apply the perturbation optimized on the source model to the target model, for each metric of prediction error, we calculate the ratio of prediction error on the target model to prediction error on the source model. Finally, we take the average of six ratio numbers corresponding to six metrics as the final score.

The results are shown in Table~\ref{tab:transfer}.
% First, all transferability scores are below 1 which means the perturbation on target models is less effective than on the source models. The optimization of perturbation indeed exploits specific patterns of each model that cannot apply to other models.
First of all, \textbf{90.25\%} of transferred adversarial trajectories successfully increase the prediction error compared with non-attack cases.
It shows that the perturbation exploits common internal patterns of general prediction models.
Therefore, the perturbation optimized on one model can launch attacks on other models as well.
It means the perturbation brings common patterns causing deviation of the predicted trajectory.
Second, the transferability is highly correlated with the target model: transferring the perturbation to \emph{FQA} is easier but to \emph{Trajectron++} is harder. We hypothesize that it is because \emph{Trajectron++} leverages more features of trajectories so that the perturbation optimized on fewer features cannot completely reproduce the high prediction error on \emph{Trajectron++}. 
% \jc{the main message in this paragraph is not clear. You are undermining your arguments. We actually want our attack to have some transferability so we can say that even with traces optimized on other models, the attacks can still achieve substantial errors than random perturbation, showing some intenral patterns/features shared in these models. We argue this is a serious problem ...}

\begin{figure*}[t]
\centering
\includegraphics[width=0.80\textwidth]{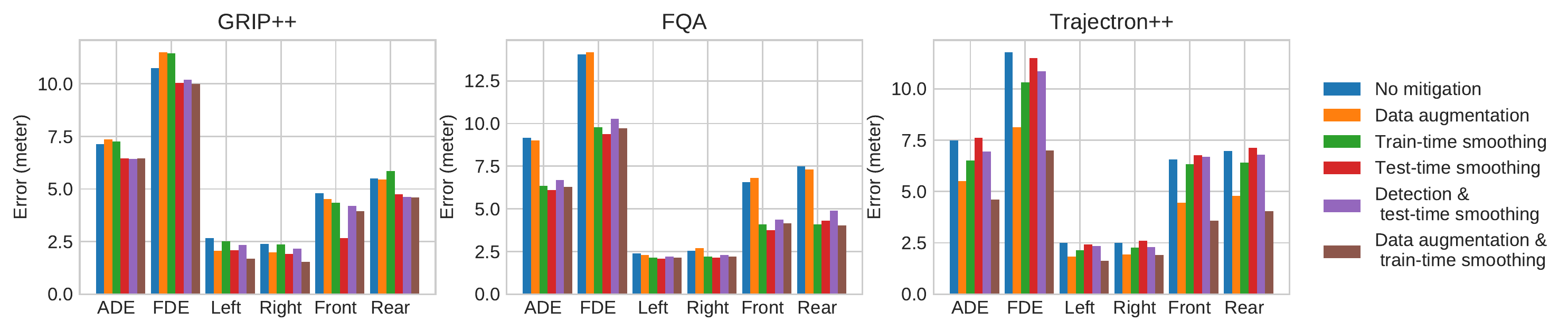}
\caption{Prediction error (single-frame, \emph{Apolloscape} dataset) before and after various mitigation methods.}
\label{fig:defense}
\end{figure*}

\begin{figure}[t]
\centering
\includegraphics[width=0.48\textwidth,height=10em]{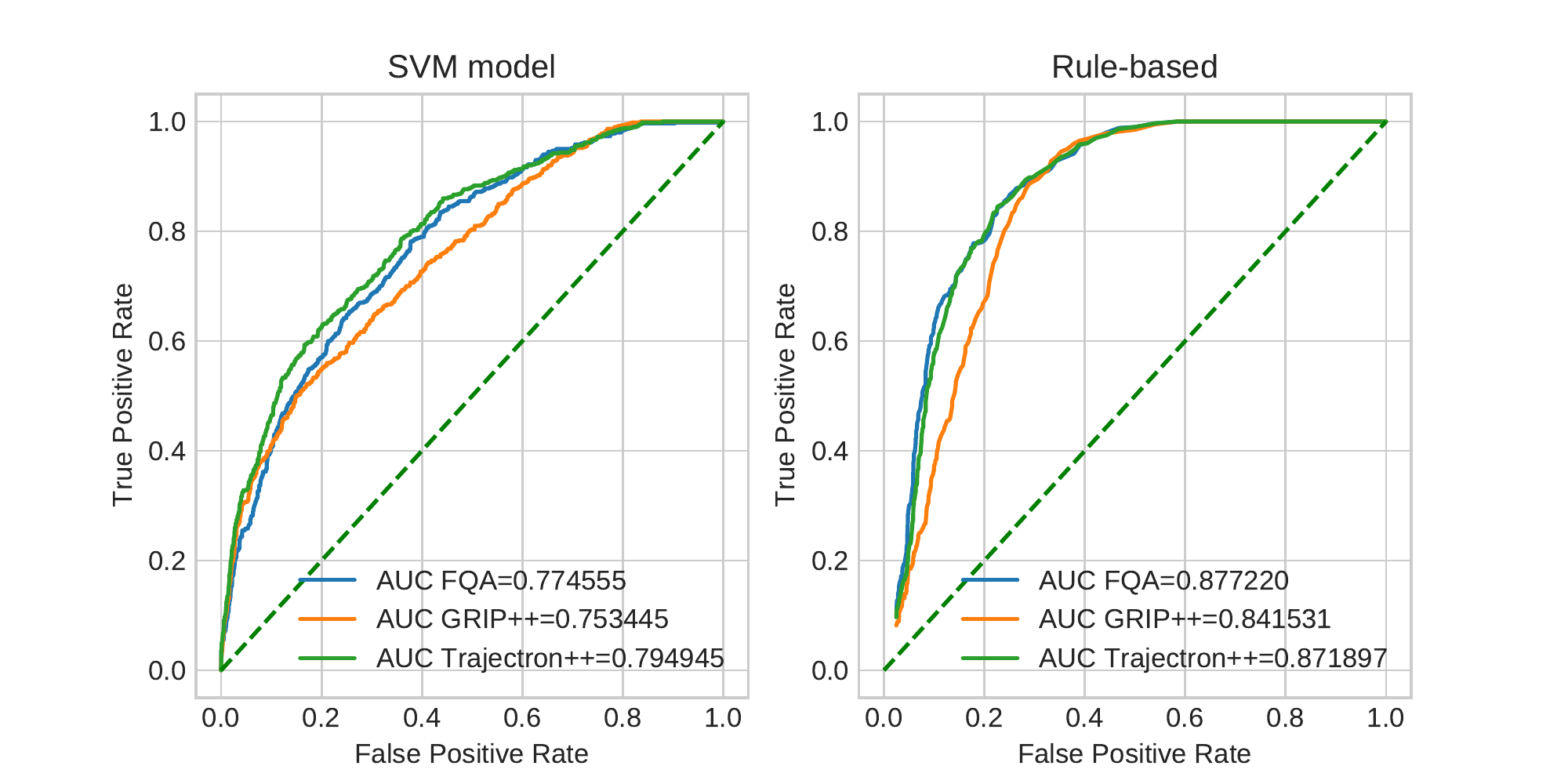}
\caption{ROC of detection of adversarial trajectories (\emph{Apolloscape} dataset, 3-second trajectories).}
\label{fig:detect}
\end{figure}

\subsection{Mitigation Results}
\label{sec:eval-defense}

We test the mitigation mechanisms using the three models and \emph{Apolloscape} dataset. The result is shown in Figure~\ref{fig:defense}. We assume that the attacker has full knowledge of the mitigation method and applies the same mitigation method on each prediction during the white-box attack. Our convolution-based trajectory smoothing is differentiable so that the computed gradient can directly involve the effect of the mitigation. If the smoothing method is replaced by non-differentiable ones, the attacker can also approximate the gradient using a differentiable function given the full knowledge of the smoothing algorithm.

\noindent
\textbf{Train-time mitigation}.
The effectiveness of data augmentation and trajectory smoothing varies on different models.
Data augmentation is effective on \emph{Trajectron++} (prediction error reduced by 24\%), which deploys a complicated network structure on low-dimensional data (i.e., trajectories). Data augmentation can alleviate the over-fitting issues of \emph{Trajectron++}.
Trajectory smoothing is effective on \emph{FQA}. \emph{FQA} model has potential under-fitting issues since its predicted trajectories mainly depend on the direction of velocity at the last time frame and have a high error on curve trajectories. Trajectory smoothing cannot solve the problem of the model but directly alleviate the impact of adversarial perturbation on the last two observed trajectory positions.
If we apply data augmentation and trajectory smoothing at train time simultaneously, the prediction error under attacks is reduced by 26\% on average while the prediction error of normal cases is increased by 11\%.

\noindent
\textbf{Test-time mitigation}.
If applying trajectory smoothing on all trajectories, the prediction error under attacks is dropped by 13\% but the normal-case prediction error raised significantly by 28\%. This is because the testing data distribution is altered to be different from the training dataset.
To address the problem, we require the detection method to distinguish adversarial trajectories from normal ones and apply smoothing only on adversarial examples. 
% \jc{discuss the gradient obfscation defense here. Basically we can say that even though it leverages gradient masking, but since we care more on the physical-world impacts also with physical constraints, even gradient masking could greatly limit the attackers' capability}

Figure~\ref{fig:detect} shows the ROC curve of two detectors mentioned in \S\ref{sec:defense}.
First, our rule-based method (i.e., the threshold on the variance of acceleration) has better performance than SVM classification in terms of true positive rate (TPR) and false positive rate (FPR). This result confirms that the variance of acceleration over time is the key difference between adversarial and normal trajectories.
Second, the detection of the adversarial trajectories for the three models has similar accuracy. It proves the generality of our detection method across various models. 
Finally, we deploy the rule-based method which has 88\% TPR and 27\% FPR.
By integrating the detection, the test-time smoothing reduces the prediction error of adversarial cases by 12\% while raising the prediction error of normal cases by only 6\%.

\noindent
\textbf{Limitation of mitigation}.
As mentioned in \S\ref{sec:eval-attack}, some normal trajectories also have the adversarial pattern, which results in relatively high FPR of detection.
Train-time methods introduce noise in training data as they change the spatial features of the original dataset. Test-time anomaly detection suffers from high FPR thus unnecessary smoothing is applied on some normal cases.
Both approaches improve adversarial robustness at a cost of slightly worse performance in some normal cases.
A complete defense of adversarial trajectories is a promising future work.
% Therefore, the mitigation improves the performance of adversarial cases at a cost of worse performance on some normal cases.

\subsection{Case study}
\label{sec:case-study}

\begin{figure}[!t]
\centering
\includegraphics[width=0.48\textwidth]{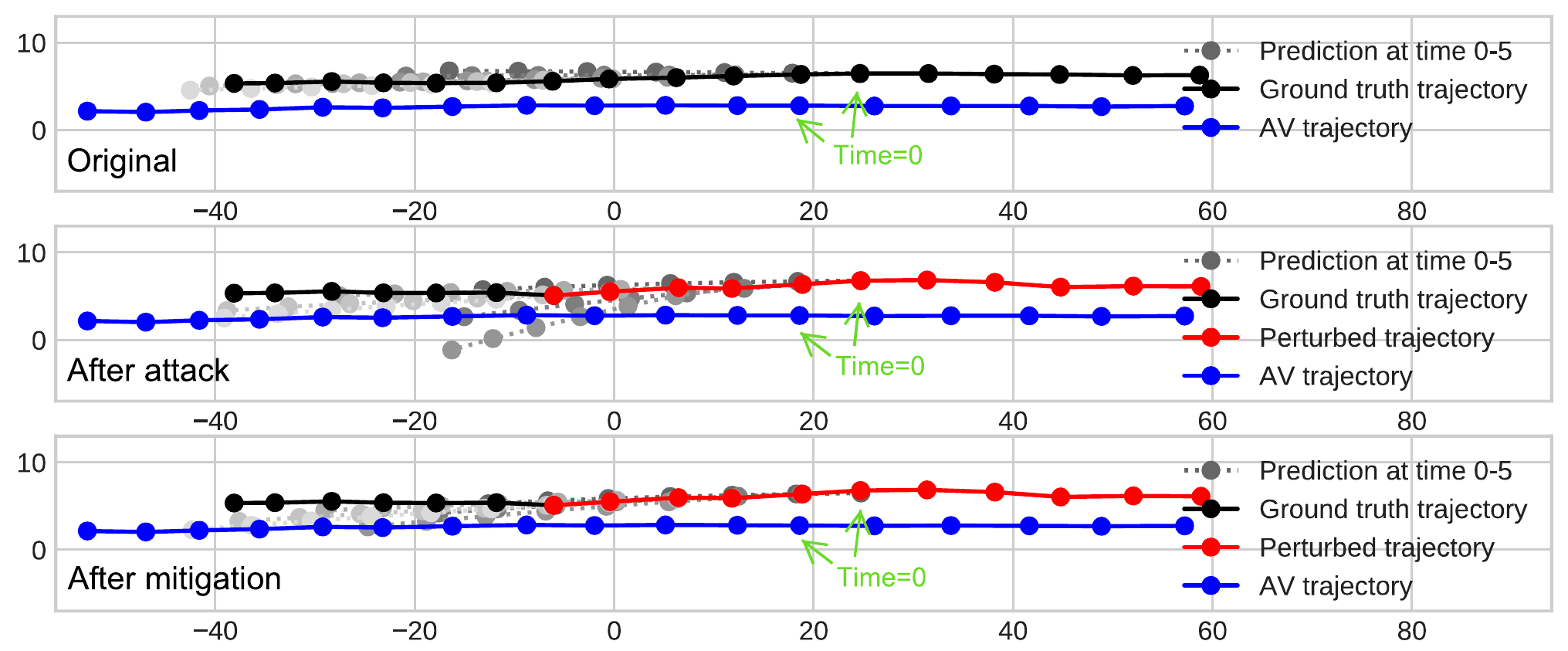}
\caption{An example of spoofing fake lane changing behavior by adversarial perturbation (\emph{GRIP++} model, \emph{Apolloscape} dataset).}
\label{fig:case-study}
\end{figure}

In this section, we demonstrate the real-world impact of the adversarial perturbation by one case study. More case studies are in the appendix.

Figure~\ref{fig:case-study} shows one scenario where the adversarial perturbation spoofs a fake lane change and causes a hard brake of the AV. 
In this scenario, the other vehicle (OV) is driving alongside the AV (we omit other objects for demonstration) and the prediction is accurate in this case (lateral deviation is 0.18 meters).
After perturbation (deviation bound of 0.5 meters, 3-second length, maximizing deviation to left), the average deviation to left is significantly increased to 1.27 meters (7$\times$).
What is worse, the high error directly affects the decision making of the AV.
At time frame 0-2, the predicted OV's trajectory crosses with the AV's future trajectory, looking like a lane changing behavior.
According to AV planning logic (e.g., open-source planning code of Baidu Apollo~\cite{apollo}), the AV will try to stop behind the cross point to yield the OV and the deceleration reaches 12 $m/s^2$, which exceeds the maximum deceleration of normal driving configured in AV software. Such a hard brake substantially increases the risk of rear-end collisions.
After applying the train-time mitigation, the deviation to left is reduced to 0.91 meters. Though the predicted trajectory and AV's future trajectory cross, the AV only needs a deceleration of 6 $m/s^2$.

We successfully reproduce the attack on real-world AV system \emph{Baidu Apollo 6.0}~\cite{apollo}. We construct the driving scenario in \emph{LGSVL} simulator~\cite{lgsvl}. Details are in the appendix.
% In the above example, the directed deviation creates fake interaction between the OV and the AV. Similarly, the deviation of prediction may also hide real interaction and shorten the safe distance between vehicles. For instance, 

\section{Conclusion}
\label{sec:conclusion}

We present the first effort of analyzing adversarial robustness of trajectory prediction. 
% which perturbs the observed trajectories to maximize prediction error 
% and test existing prediction models against such adversary. 
From the evaluation of our proposed attack, prediction models are generally vulnerable to adversarial perturbation and may cause dangerous AV behavior such as hard brakes.
We shed light on the necessity of evaluating worst-case prediction accuracy under hard scenarios or adversarial examples.
To improve adversarial robustness of trajectory prediction, we propose several mitigation methods. We also suggest leveraging map information and semantic of driving rules to guide prediction.

\noindent
\textbf{Acknowledgments.}  This work was supported by NSF under the grant CNS-1930041, CNS-1932464, CNS-1929771, CNS-2145493 and the National AI Institute for Edge Computing Leveraging Next Generation Wireless Networks, Grant \#2112562.

% \clearpage
%%%%%%%%% REFERENCES
{\small
\bibliographystyle{ieee_fullname}
\bibliography{egbib}

\begin{thebibliography}{10}\itemsep=-1pt

\bibitem{autoware}
{{Autoware: Open-source software for self-driving vehicles}}.
\newblock \url{https://gitlab.com/autowarefoundation/autoware.ai}, 2020.

\bibitem{ngsim}
{{Traffic Analysis Tools: Next Generation Simulation}}.
\newblock \url{https://ops.fhwa.dot.gov/trafficanalysistools/ngsim.htm}, 2020.

\bibitem{apollo}
{Baidu Apollo}.
\newblock \url{http://apollo.auto}, 2021.

\bibitem{athalye2018obfuscated}
Anish Athalye, Nicholas Carlini, and David Wagner.
\newblock Obfuscated gradients give a false sense of security: Circumventing
  defenses to adversarial examples.
\newblock In {\em International conference on machine learning}, pages
  274--283. PMLR, 2018.

\bibitem{nuscenes}
Holger Caesar, Varun Bankiti, Alex~H Lang, Sourabh Vora, Venice~Erin Liong,
  Qiang Xu, Anush Krishnan, Yu Pan, Giancarlo Baldan, and Oscar Beijbom.
\newblock nuscenes: A multimodal dataset for autonomous driving.
\newblock In {\em Proceedings of the IEEE/CVF conference on computer vision and
  pattern recognition}, pages 11621--11631, 2020.

\bibitem{cao2019adversarial}
Yulong Cao, Chaowei Xiao, Benjamin Cyr, Yimeng Zhou, Won Park, Sara Rampazzi,
  Qi~Alfred Chen, Kevin Fu, and Z~Morley Mao.
\newblock Adversarial sensor attack on lidar-based perception in autonomous
  driving.
\newblock In {\em Proceedings of the 2019 ACM SIGSAC conference on computer and
  communications security}, pages 2267--2281, 2019.

\bibitem{carlini2019evaluating}
Nicholas Carlini, Anish Athalye, Nicolas Papernot, Wieland Brendel, Jonas
  Rauber, Dimitris Tsipras, Ian Goodfellow, Aleksander Madry, and Alexey
  Kurakin.
\newblock On evaluating adversarial robustness.
\newblock {\em arXiv preprint arXiv:1902.06705}, 2019.

\bibitem{chandra2019traphic}
Rohan Chandra, Uttaran Bhattacharya, Aniket Bera, and Dinesh Manocha.
\newblock Traphic: Trajectory prediction in dense and heterogeneous traffic
  using weighted interactions.
\newblock In {\em Proceedings of the IEEE/CVF Conference on Computer Vision and
  Pattern Recognition}, pages 8483--8492, 2019.

\bibitem{chandra2019robusttp}
Rohan Chandra, Uttaran Bhattacharya, Christian Roncal, Aniket Bera, and Dinesh
  Manocha.
\newblock Robusttp: End-to-end trajectory prediction for heterogeneous
  road-agents in dense traffic with noisy sensor inputs.
\newblock In {\em ACM Computer Science in Cars Symposium}, pages 1--9, 2019.

\bibitem{chandra2020forecasting}
Rohan Chandra, Tianrui Guan, Srujan Panuganti, Trisha Mittal, Uttaran
  Bhattacharya, Aniket Bera, and Dinesh Manocha.
\newblock Forecasting trajectory and behavior of road-agents using spectral
  clustering in graph-lstms.
\newblock {\em IEEE Robotics and Automation Letters}, 5(3):4882--4890, 2020.

\bibitem{choi2021shared}
Chiho Choi, Joon~Hee Choi, Jiachen Li, and Srikanth Malla.
\newblock Shared cross-modal trajectory prediction for autonomous driving.
\newblock In {\em Proceedings of the IEEE/CVF Conference on Computer Vision and
  Pattern Recognition}, pages 244--253, 2021.

\bibitem{croce2020reliable}
Francesco Croce and Matthias Hein.
\newblock Reliable evaluation of adversarial robustness with an ensemble of
  diverse parameter-free attacks.
\newblock In {\em International conference on machine learning}, pages
  2206--2216. PMLR, 2020.

\bibitem{deo2020trajectory}
Nachiket Deo and Mohan~M Trivedi.
\newblock Trajectory forecasts in unknown environments conditioned on
  grid-based plans.
\newblock {\em arXiv preprint arXiv:2001.00735}, 2020.

\bibitem{dong2020benchmarking}
Yinpeng Dong, Qi-An Fu, Xiao Yang, Tianyu Pang, Hang Su, Zihao Xiao, and Jun
  Zhu.
\newblock Benchmarking adversarial robustness on image classification.
\newblock In {\em Proceedings of the IEEE/CVF Conference on Computer Vision and
  Pattern Recognition}, pages 321--331, 2020.

\bibitem{wang2021advsim}
Jingkang~Wang et. al.
\newblock Advsim: Generating safety-critical scenarios for self-driving
  vehicles.
\newblock In {\em CVPR}, 2021.

\bibitem{apolloscape}
Xinyu Huang, Xinjing Cheng, Qichuan Geng, Binbin Cao, Dingfu Zhou, Peng Wang,
  Yuanqing Lin, and Ruigang Yang.
\newblock The apolloscape dataset for autonomous driving.
\newblock In {\em Proceedings of the IEEE Conference on Computer Vision and
  Pattern Recognition Workshops}, pages 954--960, 2018.

\bibitem{jia2020fooling}
Yunhan Jia, Yantao Lu, Junjie Shen, Qi~A Chen, Zhenyu Zhong, and Tao Wei.
\newblock Fooling detection alone is not enough: First adversarial attack
  against multiple object tracking.
\newblock In {\em International Conference on Learning Representations (ICLR)},
  2020.

\bibitem{fqa}
Nitin Kamra, Hao Zhu, Dweep Trivedi, Ming Zhang, and Yan Liu.
\newblock Multi-agent trajectory prediction with fuzzy query attention.
\newblock {\em arXiv preprint arXiv:2010.15891}, 2020.

\bibitem{kennedy1995particle}
James Kennedy and Russell Eberhart.
\newblock Particle swarm optimization.
\newblock In {\em Proceedings of ICNN'95-international conference on neural
  networks}, volume~4, pages 1942--1948. IEEE, 1995.

\bibitem{grip++}
Xin Li, Xiaowen Ying, and Mooi~Choo Chuah.
\newblock Grip++: Enhanced graph-based interaction-aware trajectory prediction
  for autonomous driving.
\newblock {\em arXiv preprint arXiv:1907.07792}, 2019.

\bibitem{ma2019trafficpredict}
Yuexin Ma, Xinge Zhu, Sibo Zhang, Ruigang Yang, Wenping Wang, and Dinesh
  Manocha.
\newblock Trafficpredict: Trajectory prediction for heterogeneous
  traffic-agents.
\newblock In {\em Proceedings of the AAAI Conference on Artificial
  Intelligence}, volume~33, pages 6120--6127, 2019.

\bibitem{madry2017towards}
Aleksander Madry, Aleksandar Makelov, Ludwig Schmidt, Dimitris Tsipras, and
  Adrian Vladu.
\newblock Towards deep learning models resistant to adversarial attacks.
\newblock {\em arXiv preprint arXiv:1706.06083}, 2017.

\bibitem{marchetti2020mantra}
Francesco Marchetti, Federico Becattini, Lorenzo Seidenari, and Alberto~Del
  Bimbo.
\newblock Mantra: Memory augmented networks for multiple trajectory prediction.
\newblock In {\em Proceedings of the IEEE/CVF Conference on Computer Vision and
  Pattern Recognition}, pages 7143--7152, 2020.

\bibitem{mohamed2020social}
Abduallah Mohamed, Kun Qian, Mohamed Elhoseiny, and Christian Claudel.
\newblock Social-stgcnn: A social spatio-temporal graph convolutional neural
  network for human trajectory prediction.
\newblock In {\em Proceedings of the IEEE/CVF Conference on Computer Vision and
  Pattern Recognition}, pages 14424--14432, 2020.

\bibitem{narayanan2021divide}
Sriram Narayanan, Ramin Moslemi, Francesco Pittaluga, Buyu Liu, and Manmohan
  Chandraker.
\newblock Divide-and-conquer for lane-aware diverse trajectory prediction.
\newblock In {\em Proceedings of the IEEE/CVF Conference on Computer Vision and
  Pattern Recognition}, pages 15799--15808, 2021.

\bibitem{pedregosa2011scikit}
Fabian Pedregosa, Ga{\"e}l Varoquaux, Alexandre Gramfort, Vincent Michel,
  Bertrand Thirion, Olivier Grisel, Mathieu Blondel, Peter Prettenhofer, Ron
  Weiss, Vincent Dubourg, et~al.
\newblock Scikit-learn: Machine learning in python.
\newblock {\em the Journal of machine Learning research}, 12:2825--2830, 2011.

\bibitem{phan2020covernet}
Tung Phan-Minh, Elena~Corina Grigore, Freddy~A Boulton, Oscar Beijbom, and
  Eric~M Wolff.
\newblock Covernet: Multimodal behavior prediction using trajectory sets.
\newblock In {\em Proceedings of the IEEE/CVF Conference on Computer Vision and
  Pattern Recognition}, pages 14074--14083, 2020.

\bibitem{lgsvl}
Guodong Rong, Byung~Hyun Shin, Hadi Tabatabaee, Qiang Lu, Steve Lemke,
  M{\=a}rti{\c{n}}{\v{s}} Mo{\v{z}}eiko, Eric Boise, Geehoon Uhm, Mark Gerow,
  Shalin Mehta, et~al.
\newblock Lgsvl simulator: A high fidelity simulator for autonomous driving.
\newblock In {\em 2020 IEEE 23rd International conference on intelligent
  transportation systems (ITSC)}, pages 1--6. IEEE, 2020.

\bibitem{saadatnejad2021socially}
Saeed Saadatnejad, Mohammadhossein Bahari, Pedram Khorsandi, Mohammad Saneian,
  Seyed-Mohsen Moosavi-Dezfooli, and Alexandre Alahi.
\newblock Are socially-aware trajectory prediction models really
  socially-aware?
\newblock {\em arXiv preprint arXiv:2108.10879}, 2021.

\bibitem{trajectron++}
Tim Salzmann, Boris Ivanovic, Punarjay Chakravarty, and Marco Pavone.
\newblock Trajectron++: Dynamically-feasible trajectory forecasting with
  heterogeneous data.
\newblock In {\em Computer Vision--ECCV 2020: 16th European Conference,
  Glasgow, UK, August 23--28, 2020, Proceedings, Part XVIII 16}, pages
  683--700. Springer, 2020.

\bibitem{sato2021dirty}
Takami Sato, Junjie Shen, Ningfei Wang, Yunhan Jia, Xue Lin, and Qi~Alfred
  Chen.
\newblock Dirty road can attack: Security of deep learning based automated lane
  centering under physical-world attack.
\newblock In {\em Proceedings of the 29th USENIX Security Symposium (USENIX
  Security’21)}, 2021.

\bibitem{shafiee2021introvert}
Nasim Shafiee, Taskin Padir, and Ehsan Elhamifar.
\newblock Introvert: Human trajectory prediction via conditional 3d attention.
\newblock In {\em Proceedings of the IEEE/CVF Conference on Computer Vision and
  Pattern Recognition}, pages 16815--16825, 2021.

\bibitem{shi2021sgcn}
Liushuai Shi, Le Wang, Chengjiang Long, Sanping Zhou, Mo Zhou, Zhenxing Niu,
  and Gang Hua.
\newblock Sgcn: Sparse graph convolution network for pedestrian trajectory
  prediction.
\newblock In {\em Proceedings of the IEEE/CVF Conference on Computer Vision and
  Pattern Recognition}, pages 8994--9003, 2021.

\bibitem{steinwart2008support}
Ingo Steinwart and Andreas Christmann.
\newblock {\em Support vector machines}.
\newblock Springer Science \& Business Media, 2008.

\bibitem{sun2020reciprocal}
Hao Sun, Zhiqun Zhao, and Zhihai He.
\newblock Reciprocal learning networks for human trajectory prediction.
\newblock In {\em Proceedings of the IEEE/CVF Conference on Computer Vision and
  Pattern Recognition}, pages 7416--7425, 2020.

\bibitem{sun2020towards}
Jiachen Sun, Yulong Cao, Qi~Alfred Chen, and Z~Morley Mao.
\newblock Towards robust lidar-based perception in autonomous driving: General
  black-box adversarial sensor attack and countermeasures.
\newblock In {\em 29th $\{$USENIX$\}$ Security Symposium ($\{$USENIX$\}$
  Security 20)}, pages 877--894, 2020.

\bibitem{sun2021adversarially}
Jiachen Sun, Yulong Cao, Christopher~B Choy, Zhiding Yu, Anima Anandkumar,
  Zhuoqing~Morley Mao, and Chaowei Xiao.
\newblock Adversarially robust 3d point cloud recognition using
  self-supervisions.
\newblock {\em Advances in Neural Information Processing Systems}, 34, 2021.

\bibitem{sun2020recursive}
Jianhua Sun, Qinhong Jiang, and Cewu Lu.
\newblock Recursive social behavior graph for trajectory prediction.
\newblock In {\em Proceedings of the IEEE/CVF Conference on Computer Vision and
  Pattern Recognition}, pages 660--669, 2020.

\bibitem{sun2020adversarial}
Jiachen Sun, Karl Koenig, Yulong Cao, Qi~Alfred Chen, and Z~Morley Mao.
\newblock On adversarial robustness of 3d point cloud classification under
  adaptive attacks.
\newblock {\em arXiv preprint arXiv:2011.11922}, 2020.

\bibitem{wang2021human}
Jiahang Wang, Sheng Jin, Wentao Liu, Weizhong Liu, Chen Qian, and Ping Luo.
\newblock When human pose estimation meets robustness: Adversarial algorithms
  and benchmarks.
\newblock In {\em Proceedings of the IEEE/CVF Conference on Computer Vision and
  Pattern Recognition}, pages 11855--11864, 2021.

\bibitem{wang2021f}
Jue Wang, Ping Wang, Chao Zhang, Kuifeng Su, and Jun Li.
\newblock F-net: Fusion neural network for vehicle trajectory prediction in
  autonomous driving.
\newblock In {\em ICASSP 2021-2021 IEEE International Conference on Acoustics,
  Speech and Signal Processing (ICASSP)}, pages 4095--4099. IEEE, 2021.

\bibitem{yuan2021agentformer}
Ye Yuan, Xinshuo Weng, Yanglan Ou, and Kris Kitani.
\newblock Agentformer: Agent-aware transformers for socio-temporal multi-agent
  forecasting.
\newblock {\em arXiv preprint arXiv:2103.14023}, 2021.

\end{thebibliography}
}

\newpage
\appendix

\section{Implementation Details}

\subsection{Open-sourced Code}

Our implementation is open source at Github (\url{https://github.com/zqzqz/AdvTrajectoryPrediction}), which is a framework of testing adversarial attacks and mitigation methods on trajectory prediction algorithms.

\subsection{Bounds of Physical Properties}

\begin{table}[t]
\scriptsize
\caption{Maximum bounds of physical properties.}
  \label{tab:bounds}
  \centering
  \begin{tabular}{| c | c | c | c | c | c |}
    \noalign{\global\arrayrulewidth1pt}\hline\noalign{\global\arrayrulewidth0.4pt}
    Dataset & $|v|$ & $a_l$ & $a_s$ & $d a_l / dt$ & $d a_s / dt$ \\
    \noalign{\global\arrayrulewidth1pt}\hline\noalign{\global\arrayrulewidth0.4pt}
    Apolloscape~\cite{apolloscape} & 21.078 & 9.914 & 1.912 & 16.836 & 3.154 \\
    \hline
    NGSIM~\cite{ngsim} & 20.830 & 1.455 & 0.620 & 1.955 & 0.925 \\
    \hline
    nuScenes~\cite{nuscenes} & 17.198 & 2.550 & 0.936 & 3.914 & 1.070 \\
    \noalign{\global\arrayrulewidth1pt}\hline\noalign{\global\arrayrulewidth0.4pt}
  \end{tabular}
  \\\textbf{Notes}: $|v|$ -- scalar velocity ($m/s$); $a_l$ -- longitudinal acceleration ($m/s^2$); $a_s$ -- lateral acceleration ($m/s^2$); $d a_l / dt$ -- derivative of longitudinal acceleration ($m/s^3$); $d a_s / dt$ -- derivative of lateral acceleration ($m/s^3$).
\end{table}

We restrict 5 physical properties of perturbed trajectories: (1) scalar velocity, (2) longitudinal acceleration, (3) lateral acceleration, (4) derivative of longitudinal acceleration, and (5) derivative of lateral acceleration. For the three datasets (i.e., Apolloscape, NGSIM, and nuScenes), we select different bounds of physical properties according to the data distribution following the approach mentioned in the main paper. We report the values of such bounds in Table~\ref{tab:bounds}.

\subsection{Hyper-parameters of Attacks}

In general, both white-box and black-box attacks are iteration-based methods, which require parameters about evolution speed and maximum iterations.
When tuning the parameters, we monitor the objective loss over time. The parameters are proper if the loss is overall decreasing and stays low stably in the end.

For the PGD-based white box attack, we use Adam optimizer with a learning rate of 0.01 and set the maximum iteration to 100.
For PSO-based black box attack, we set the number of particles to 10, inertia weight to 1.0, acceleration coefficients to (0.5, 0.3), and the maximum iteration to 100. 
% The drop of loss of two attack approaches is demonstrated in Figure~\ref{fig:loss-drop}, respectively.
% We observe that the loss converges to a low level in the end.

\subsection{Implementation of Mitigation}

For data augmentation, we randomly select 50\% (the parameter is fine-tuned) of trajectories to add perturbation during the training. The added perturbation is random and under the hard constraints of perturbation. We also double the maximum training iteration.

For trajectory smoothing, we use a linear smoother with a convolution kernel $(\frac{1}{3},\frac{1}{3},\frac{1}{3})$, which takes the mean of three trajectory positions as the smoothed position at the middle time frame.
Formally, we denote the trajectory as ${s_1, \dots, s_i, \dots, s_N}$ where $N$ is the total length of the trajectory and $s_i$ ($i \in 1\dots N$) is the two-dimensional trajectory location at time frame $i$.
The smoothed trajectory locations $s_i' = \frac{1}{3} s_{i-1} + \frac{1}{3} s_{i} + \frac{1}{3} s_{i+1}$ ($i\in 1\dots N-1$).

For detection, we prepare a set of normal trajectories (other than training/testing data used in the attack experiments) and generate a set of abnormal trajectories by adding random perturbation on normal trajectories. We use the normal and perturbed trajectories together to fit the SVM model and find out proper thresholds. 
The SVM model uses RBF kernel, implemented in the \emph{scikit-learn} library.
The threshold parameters for detection are selected to achieve the best Area Under Curve (AUC) score of Receiver Operating Characteristics (ROC) curve.  

\section{Hard Scenarios for Prediction}
\label{sec:hard-scenarios}

\begin{figure}[t]
\centering
\includegraphics[width=0.40\textwidth]{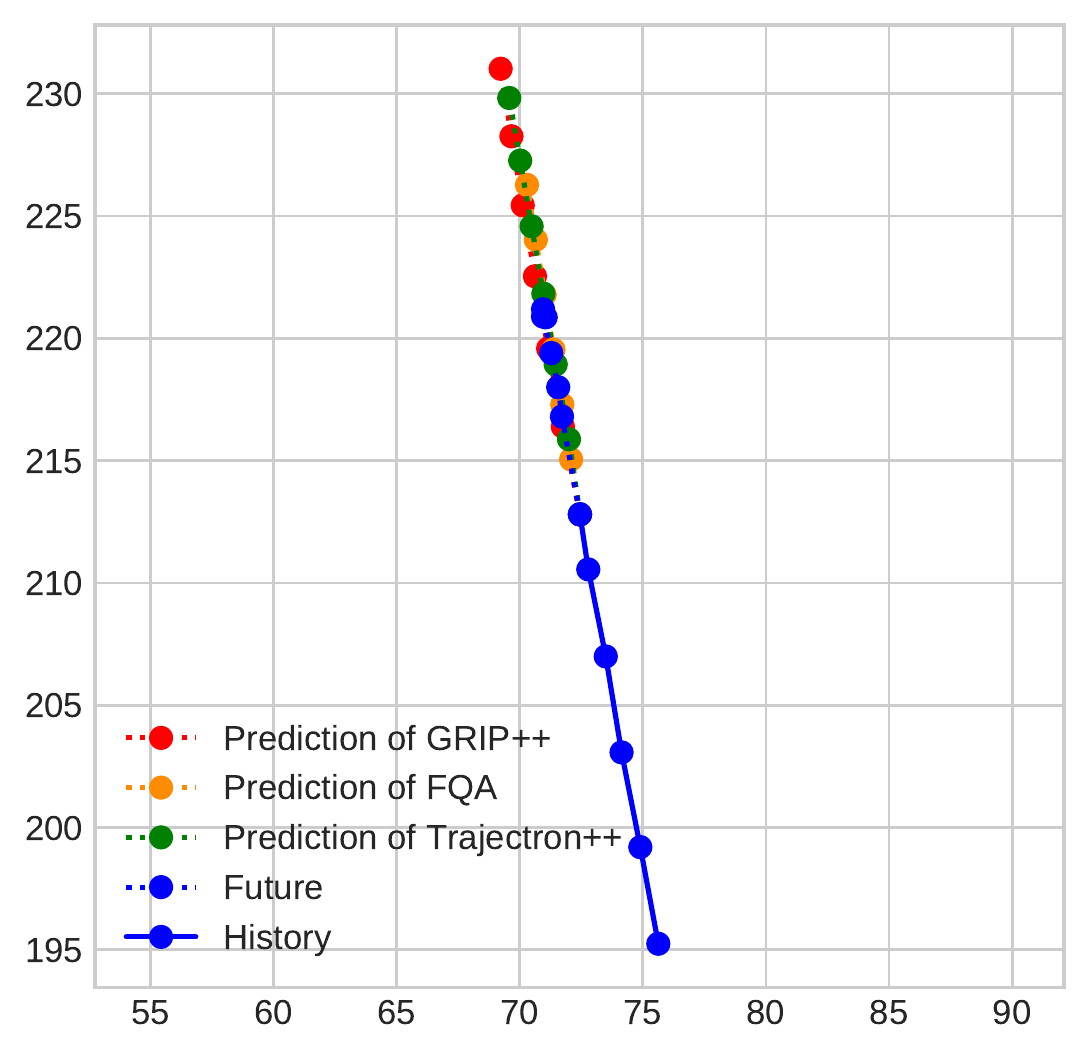}
\caption{Hard scenario: stopping at a stop sign.}
\label{fig:hard-scenario-0}
\end{figure}

\begin{figure}[t]
\centering
\includegraphics[width=0.40\textwidth]{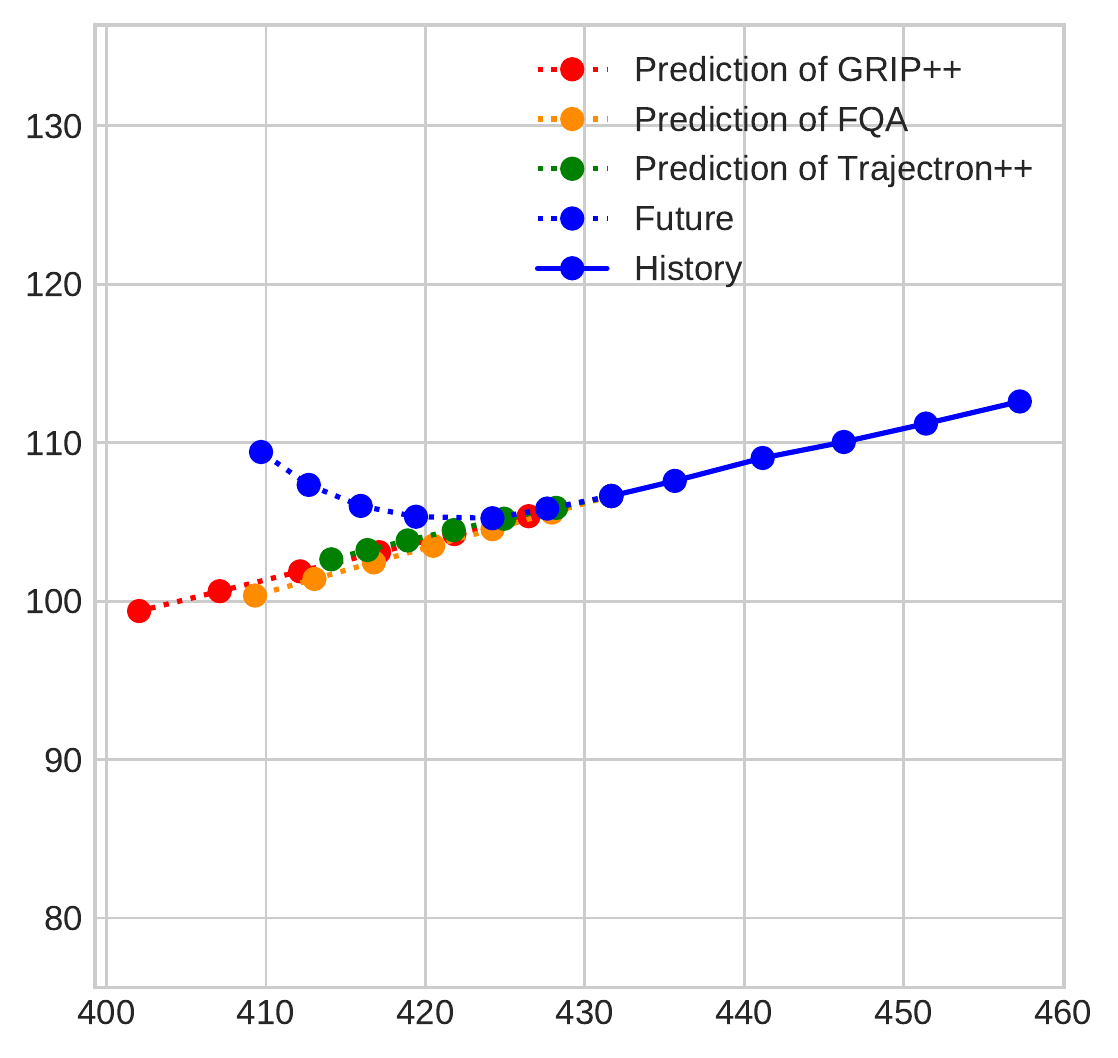}
\caption{Hard scenario: turning right at an intersection.}
\label{fig:hard-scenario-1}
\end{figure}

In the main paper we mention that existing prediction models have high prediction error on several hard scenarios. In this section, we visualize two scenarios in Figure~\ref{fig:hard-scenario-0} and Figure~\ref{fig:hard-scenario-1}. In each figure, we present the ground-truth trajectory of the target vehicle and prediction made by tested models. 
In Figure~\ref{fig:hard-scenario-0}, the target vehicle takes brake in its future trajectory and finally stops at the stop sign. However, the prediction models cannot foresee such braking behavior and the predicted trajectory still has a steady velocity.
In Figure~\ref{fig:hard-scenario-1}, the target vehicle turns right in its future trajectory but the pattern of turning is not obvious in the history trajectory. Therefore, it is hard for prediction models to correctly predict such turning trajectory.

\section{More Case Studies}
\label{sec:more-case-studies}

\begin{figure}[t]
\centering
\includegraphics[width=0.48\textwidth]{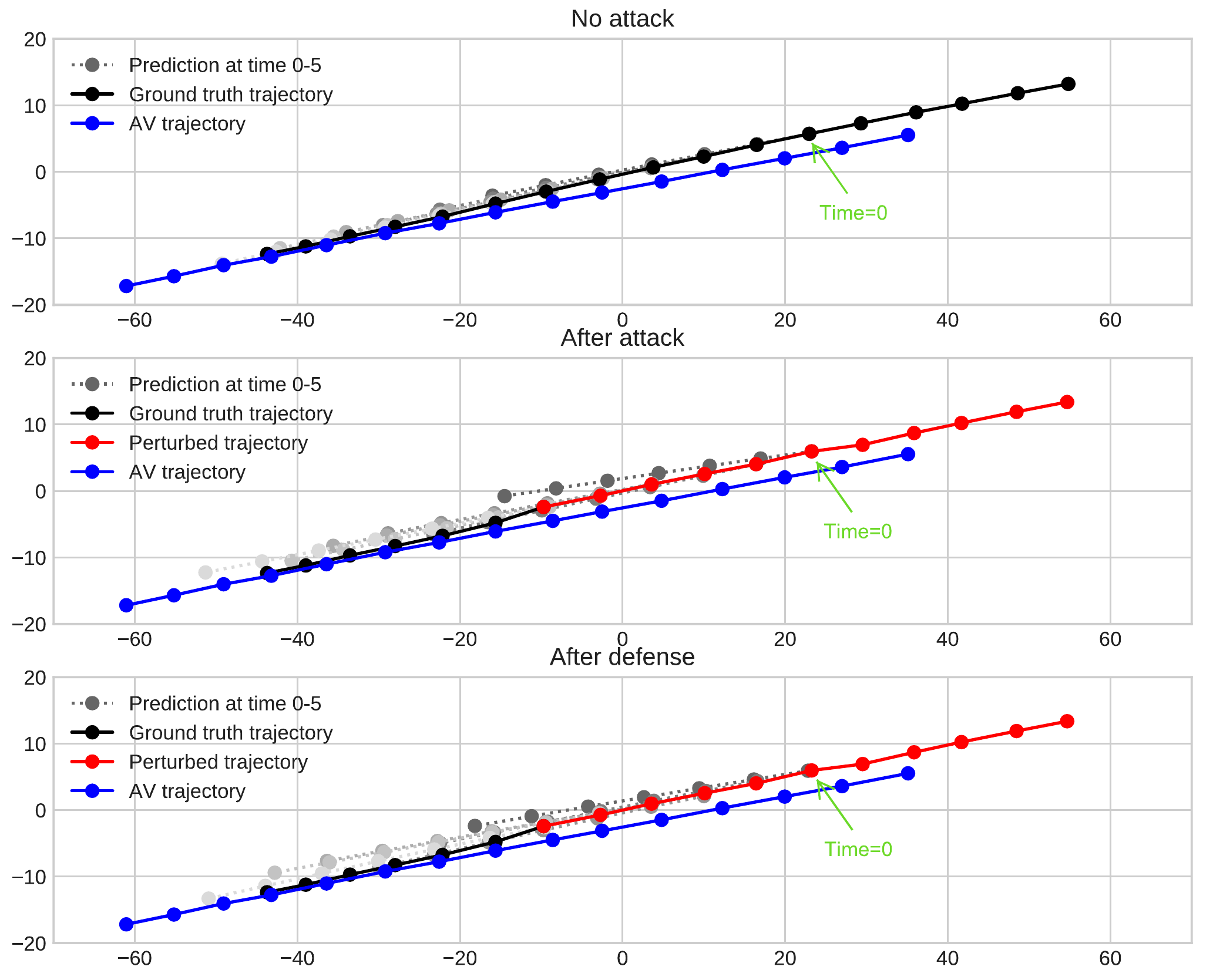}
\caption{Case study: deviation to right hides a real lane changing behavior. GRIP++ model, Apolloscape dataset, white-box multi-frame attack, mitigation of data augmentation and train-time smoothing.}
\label{fig:case-study-2}
\end{figure}

\begin{figure}[t]
\centering
\includegraphics[width=0.48\textwidth]{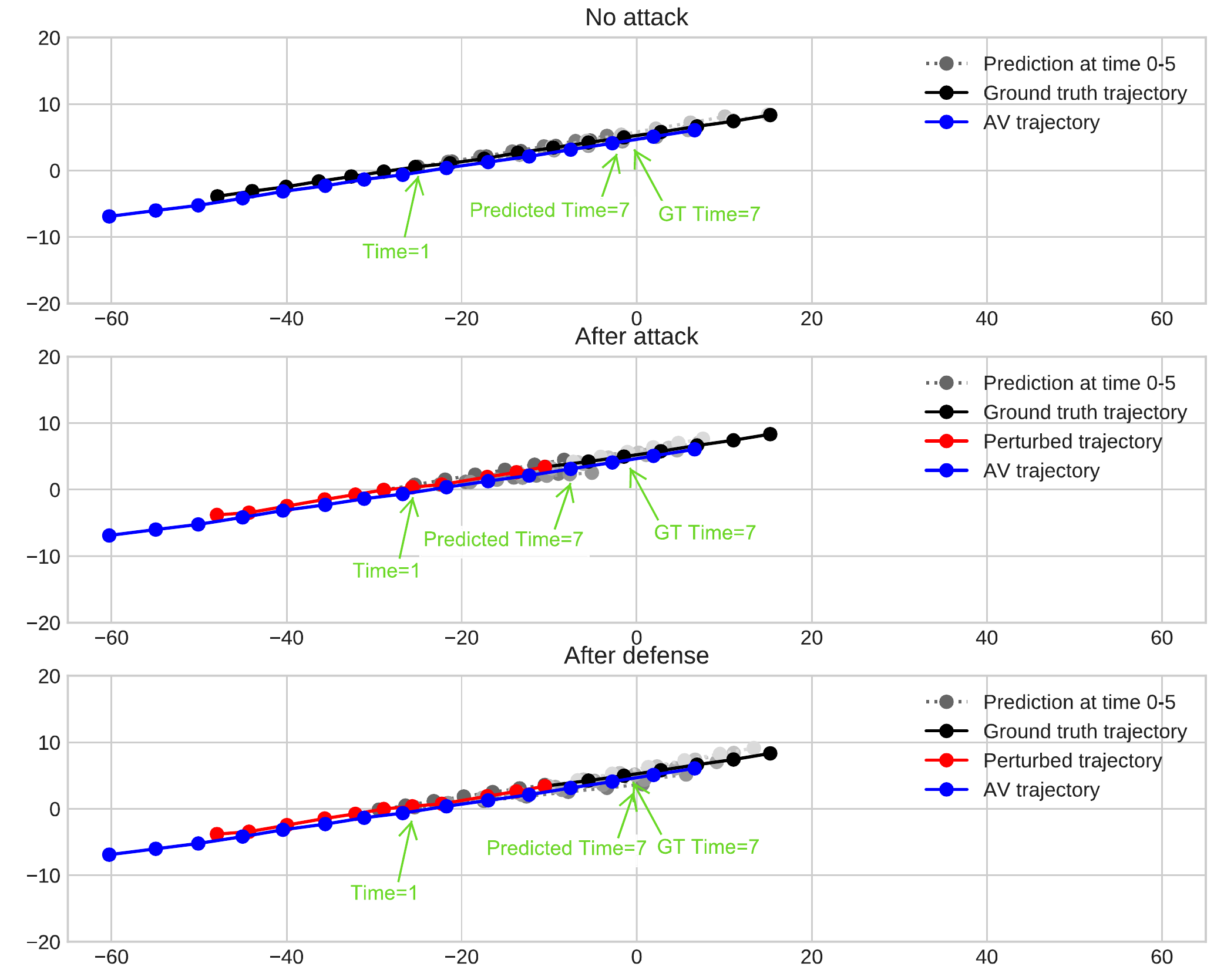}
\caption{Case study: deviation to rear spoofs a fake braking behavior. GRIP++ model, Apolloscape dataset, white-box multi-frame attack, mitigation of data augmentation and train-time smoothing.}
\label{fig:case-study-1}
\end{figure}

The case study in the main paper discusses a scenario where the attacker leverages high prediction error to spoof a fake lane changing behavior. In this section, we show two more case studies to demonstrate attack impacts considering various attack goals.

First, the adversarial trajectories with high prediction error can also hides existing driving behaviors instead of spoofing behaviors.
As shown in Figure~\ref{fig:case-study-2}, the OV is in fact shifting to the left lane (the AV's lane) and the prediction correctly captures the behavior without perturbation. The average deviation to right is 0.61 meter originally. Under the white-box attack (3-second length, 1-meter deviation bound, maximizing deviation to right), the adversarial trajectory is still natural but the predicted trajectories are going straight without any pattern of changing lanes. The average deviation to right is increased to 1.22 meters (2$\times$).
The attack significantly reduces the time for the AV to safely respond to the lane changing behavior. Originally, the AV learn the OV will change lane at time frame 1 because the prediction at time frame 2 already cross with the AV's future trajectory. After the attack, the AV does not acknowledge the lane changing until time frame 7. Hence, the AV receives the lane changing signal 2.5 seconds late which may cause a hard brake or even a rear-end collision.
What is worse, the mitigation is not effective on this case because the adversarial trajectory is smooth and natural between time frame 1 to 6.

Second, the longitude deviation is as dangerous as the lateral deviation.
In Figure~\ref{fig:case-study-1}, the OV is driving in front of the AV with a distance of about 10 meters. Originally, the OV moves in a almost constant velocity and the average deviation to rear direction is 0.22 meter. After the white-box attack (3-second length, 1-meter deviation bound, maximizing deviation to rear), the OV stays in its route but the velocity is not stable. On the adversarial trajectory, the predicted trajectories show that OV is going to decelerate and the deviation to rear is increased to 2.99 meters (14$\times$). In time frame 2 for instance, the length of the predicted trajectory is shortened by 50\% and the AV's planning logic decides to deceleration to avoid a potential collision.
Depending on the distance between the OV and the AV, the attack may cause speed drop, a hard brake, or collision of the AV.
Mitigation of trajectory smoothing is effective on this case. The trajectory smoothing alleviates the fluctuation of the OV's velocity.

\section{Realistic Attack Setting}
\label{sec:eval-realistic}

As discussed in the threat model, the adversary does not have the ground truth of other vehicles/pedestrians when computing the adversarial trajectory. Therefore, we consider a more realistic setting where the adversary includes predicted trajectories in the input of generating adversarial examples.
We select 75 scenarios in \emph{Apolloscape} dataset and analyze attacks on \emph{GRIP++} model. In each scenario, the adversary has full knowledge of trajectories in the first 3 seconds, predicts trajectories of all other objects between 3-6 seconds using \emph{GRIP++}, and generates the adversary trajectory between 3-6 seconds. Prediction error (six metrics) of the above attack is 6.82/10.41/2.61/2.38/3.92/5.53 meters, which is 95\% of the attacks using ground-truth object trajectories. The adversary can approximate the future knowledge to generate attacks whose effectiveness is almost equal to ideal white-box attacks in Table~3 in the main paper because the target vehicle's trajectory itself dominates the prediction results.

\section{Attack Real-world AV System}

\begin{figure}[!t]
    \centering
    \begin{subfigure}{0.22\textwidth}
        \centering
        \includegraphics[height=2cm]{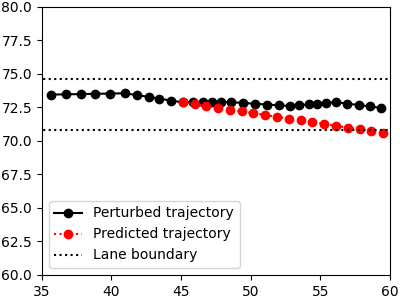}
        \caption{Attacker's perturbed trajectory and prediction made by Apollo.}
    \end{subfigure}
    \begin{subfigure}{0.22\textwidth}
        \centering
        \includegraphics[height=2cm]{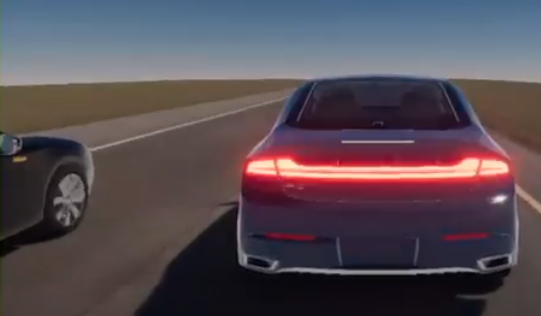}
        \caption{Simulator LGSVL's view.}
    \end{subfigure}
\caption{Reproducing a prediction attack on Baidu Apollo 6.0.}
\label{fig:apollo}
\end{figure}

Our attack is effective on real AV software, Baidu Apollo 6.0~\cite{apollo}, which uses an LSTM predictor on 2 seconds of history trajectory in a frequency of 10 Hz.
% The figure below shows that a black-box adversarial trajectory spoofs fake lane changing and makes Apollo AV brake (in LGSVL simulator), similar to Fig.7.
Figure~\ref{fig:apollo} shows that the adversarial trajectory spoofs fake lane changing (in the left figure), resulting in a brake of the right Apollo AV (in LGSVL simulator~\cite{lgsvl}).
Second, history length and frequency are system-specific settings, and prediction models mostly choose 2-3 seconds of history and 2-10 Hz. Although we showed different values for the two parameters in Tab.1, we did not observe a clear correlation between the parameters and prediction accuracy. We can openly discuss this question as future work.

% % \clearpage
% %%%%%%%%% REFERENCES
% {\small
% \bibliographystyle{ieee_fullname}
% \bibliography{egbib}
% }

% \end{document}

\end{document}